\documentclass[10pt,twocolumn,letterpaper]{article}
\usepackage{titling}
\usepackage[pagenumbers]{wacv} % To force page numbers, e.g. for an arXiv version

\usepackage{graphicx}
\usepackage{amsmath}
\usepackage{amssymb}
\usepackage{booktabs}
\usepackage{float}
\usepackage{xcolor}
\usepackage[pagebackref,breaklinks,colorlinks]{hyperref}
\usepackage[font=footnotesize,labelfont=bf]{caption} % changing the caption size for more space
\usepackage[accsupp]{axessibility}  % Improves PDF readability for those with disabilities.
\usepackage[capitalize]{cleveref}
\crefname{section}{Sec.}{Secs.}
\Crefname{section}{Section}{Sections}
\Crefname{table}{Table}{Tables}
\crefname{table}{Tab.}{Tabs.}

\def\wacvPaperID{0669} % *** Enter the WACV Paper ID here
\def\confName{WACV}
\def\confYear{2024}
\newcommand{\shrink}{\def\baselinestretch{0.92}\large\normalsize} % <0.95 makes it ugly!!
\shrink
\begin{document}

\title{Shape from Shading for Robotic Manipulation}
\author{Arkadeep Narayan Chaudhury \and
Leonid Keselman \and
Christopher G. Atkeson\\
The Robotics Institute, Carnegie Mellon University\\
5000 Forbes Avenue, Pittsburgh, Pennsylvania 15213, USA\\
{\tt\small \{arkadeepnc, leonidk, cga\}@cmu.edu}
}
\maketitle

%%%%%%%%% ABSTRACT
\begin{abstract}
   Controlling illumination can generate high quality information about object surface normals and depth discontinuities at a low computational cost. In this work we demonstrate a robot workspace-scaled controlled illumination approach that generates high quality information for table top scale objects for robotic manipulation. With our low angle of incidence directional illumination approach, we can precisely capture surface normals and depth discontinuities of monochromatic Lambertian objects. We show that this approach to shape estimation is 1) valuable for general purpose grasping with a single point vacuum gripper, 2) can measure the deformation of known objects, and 3) can estimate pose of known objects and track unknown objects in the robot's workspace. 
\end{abstract}

%%%%%%%%% BODY TEXT
\section{Introduction}\label{sc:introduction}
% To ground our motivation for this work, 
Imagine the following task:  you are trying to find a tiny screw on the ground, but the color contrast between the screw and the ground is poor, or the floor has a complex texture.  You might turn to a flashlight to aid you in this task, illuminating part of the floor with a low angle of incidence, looking for shadow cues as you hunt for the screw.

Now consider the task of examining the quality of examining the quality of stucco\cite{wiki:Stucco} or putty applied to a surface; here, a low angle of incidence light (as in the first task) can highlight high spatial frequency surface discontinuities such as surface roughness and features like creases.  A directional source of light (e.g. sunlight throughout the day), meanwhile, can highlight low spatial frequency information such as the curvature of the wall. 
%Imagine the following tasks of 1) finding a tiny screw on the ground especially when the color contrast between the screw and the background is poor or the background is complex, 2) examining the quality of stucco\cite{wiki:Stucco} or putty applied to a surface.
 % , and 3) picking up an uncommon object (e.g. the object \cref{fig:pipeline_master}) where it is not obviously clear, from prior knowledge, which face to approach for a successful grasp and, additionally some face may have un-graspable texture further complicating the task. 
 %To complete the first task, one can use a flashlight and illuminate a part of the floor with a low angle of incidence and look for shadow cues. 
 %For the second task, low angle of incidence light can highlight high spatial frequency surface discontinuities such as surface roughness and features like creases, while a directional source of light (e.g. sunlight throughout the day) can highlight low spatial frequency information such as the curvature of the wall. \newline 

This work is motivated by the observation that changing the direction of illumination can often highlight object surface features and depth discontinuities on the object surface and between the object and its surroundings. These surface features often lead to shadows and illuminated patches when viewed with a camera. Shading and shadows along with an illumination model can aid in reasoning about the surface properties of an object and help us decide how to interact with it.   
 %and 3, one can move the light around the object while reasoning about the observed highlights and shadows, conditioned on the direction of illumination. Surfaces with ``unusual'' shadows and/or highlights can be ruled out and one can determine if a part of the object is graspable by essentially reasoning about the surface normals and shadows.
  \begin{figure}
    % \centering
    \begin{subfigure}[b]{0.4\textwidth}
        \centering
            \includegraphics[width=1.1\textwidth]{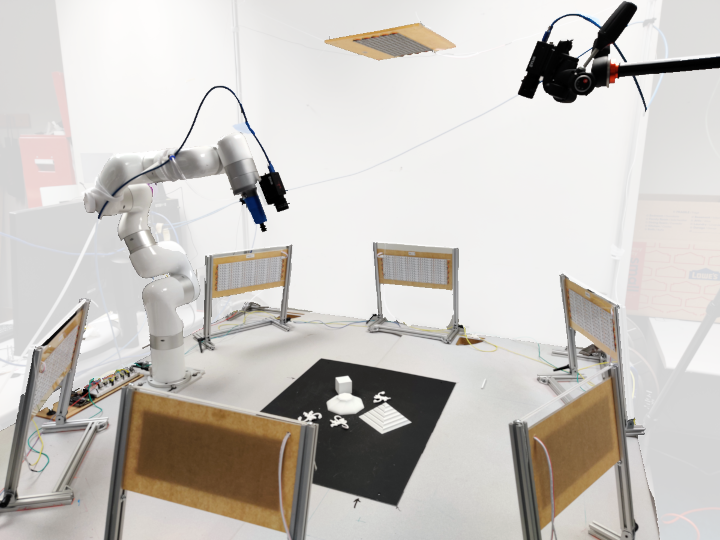}
        \subcaption{}
        \label{fig:workspace_schematic}
    \end{subfigure}
    % \\[ 2 ex] 
    \begin{subfigure}[b]{0.15\textwidth}
      \centering
      \includegraphics[width=\textwidth]{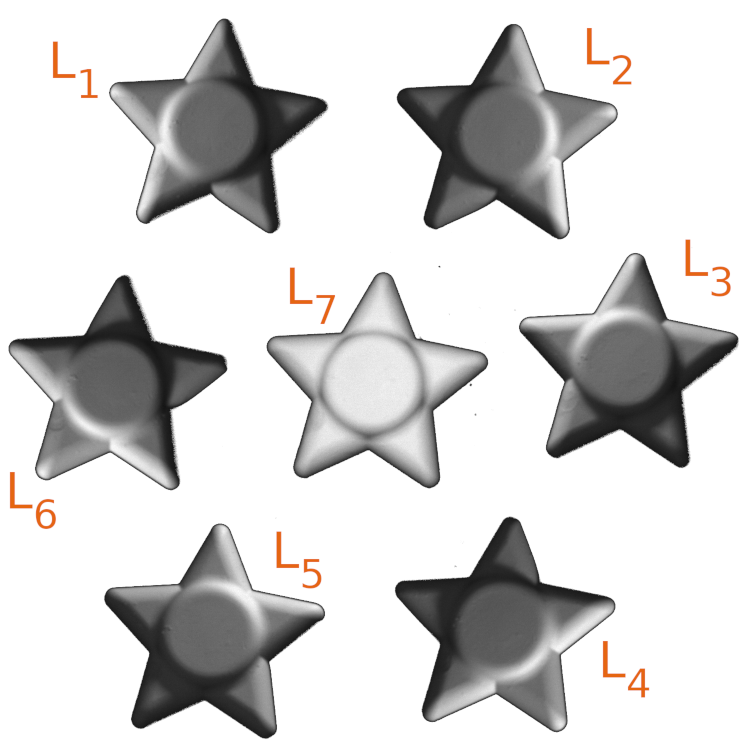}
      \caption{}
      \label{fig:light_channels_star}
    \end{subfigure} 
    \begin{subfigure}[b]{0.15\textwidth}
      \centering
      \includegraphics[width=\textwidth]{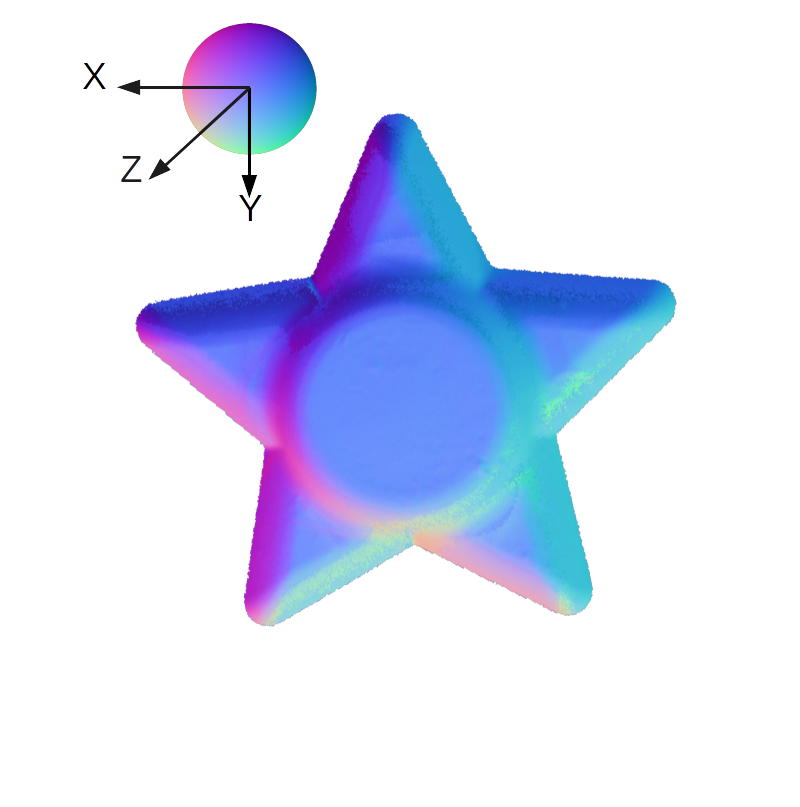}
      \caption{}
      \label{fig:normals_star}
    \end{subfigure} 
    \begin{subfigure}[b]{0.15\textwidth}
      \centering
      \includegraphics[width=\textwidth]{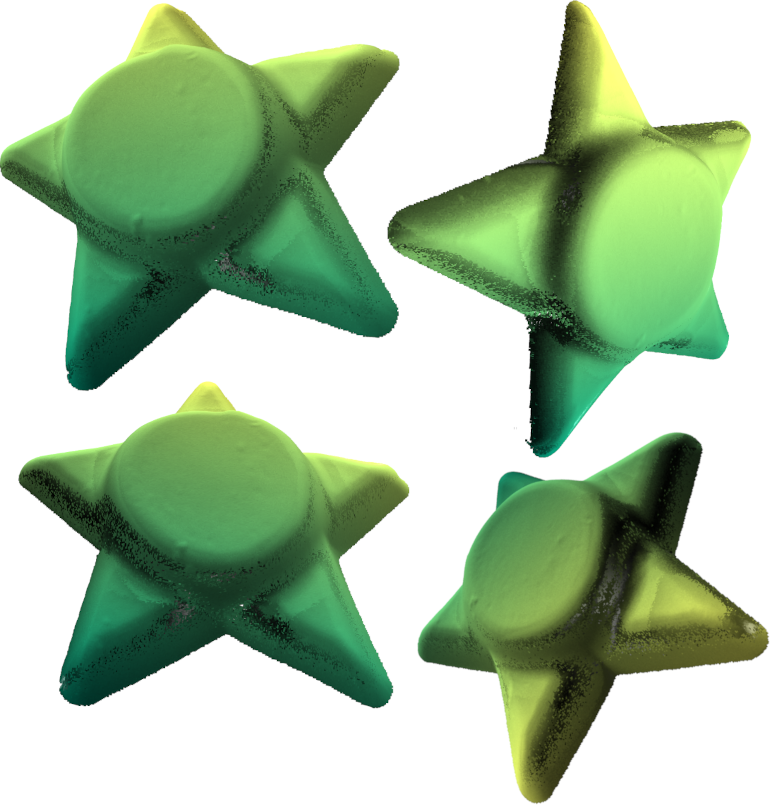}
      \caption{}
      \label{fig:pcd_star}
    \end{subfigure} 
   \caption{ \textbf{We demonstrate the value of controlled illumination approaches for robot manipulation workspaces. } Our setup (\cref{fig:workspace_schematic}) consists of 2 machine vision cameras ($C_1$ mounted to the robot and $C_2$) overlooking a robot's workspace. The illumination of the workspace is controlled using 7 directional lights -- six low angle of incidence light sources $L_{1:6}$ placed on the table and one overhead light source $L_7$. We capture seven images of the object (\cref{fig:light_channels_star}), use standard photometric stereo to calculate object surface normals (color coded in \cref{fig:normals_star}), and optionally derive a 3D representation of the object's surface (\cref{fig:pcd_star}) from calculated normals (\cref{fig:normals_star}).}
   \label{fig:opening_figure}
   \end{figure}
 Although classical techniques in multi-view geometry~\cite{Schoenberger2016Colmap}, shape from texture~\cite{Aloimonos1988shape, Verbin2020}, high performance area scanners~\cite{Photoneo,Ensenso} and tactile sensing~\cite{Yuan2017gelsight,Donlon2018gelslim} exist for solving similar problems, multi-modal sensing adds complexity to the problem by requiring the camera poses and object features to be tracked across views and sensors. Coordinating tactile sensing with vision~\cite{Chaudhury2022} adds the additional complexities of discovering correspondences between cameras and tactile sensor data and accounting for tactile sensor drift. Our experiments focus on precise and effective sensing in a robot manipulation setup ~\cite{kalashnikov2018scalable,kroemer2021review}.
 
 % \indent To that end, this paper proposes using \textit{actively controlled illumination} and tries to address simplified versions of the motivating question: In the absence of highlights due to specularities and textures, how can controlled directional illumination be used to estimate surface properties of objects to primarily assist in robotic manipulation. \newline
To that end, this paper proposes using actively controlled illumination and tries to address a simplified version of the motivating question:  How can controlled illumination be used to estimate surface properties of objects to make robotic manipulation easier?  The simplifications include painting objects with matte paint to remove highlights due to specularities, and using only one color of paint (white) to make the reflectance function uniform over the object.
 
Active control of sensing parameters is a classical topic in robotics research -- \cite{Bajcsy1988, Bajcsy2018revisiting,Aloimonos1988} define ``active perception'' as controlling environment parameters (e.g. illumination), extrinsic (e.g. sensor location) and intrinsic (e.g sensor gain, focal length) sensor parameters for better perception. Our work can be classified as a subset of active perception research where the scene illumination, camera poses, and some of the camera intrinsic parameters are actively controlled to solve the perception task at hand.
 % We divide the problem of active perception through controlled illumination into two parts -- capturing and processing images of objects, and using the processed images to control robot manipulation. 
 
 Although we will handle more general reflectance functions in future work, for this work, we assume that all objects have monochromatic Lambertian surface reflectances, i.e.\ they do not have specularities and changes in reflectances due to colored patches on the object. We enforce this assumption by coating all objects with matte white paint\footnote{we use RustOleum 7790830 Flat White spray paint} when necessary. All of our objects are single rigid bodies with no articulations and when needed, we assume that we have 3D models available to us. We show that: 
 % and segmentation masks of objects  \newline
% \indent We show that:
\begin{itemize}
    \item Controlling illumination of a robot's workspace yields high quality information about surface normals and discontinuities -- significantly better than commercially available 3D sensors for table top scale objects ($\leq$20cm$^3$).
    \item The computed normals and surface discontinuity information can aid in robot grasping tasks.
    \item A controlled illumination approach can help us estimate deformation of known objects, and estimate poses of known and unknown objects.
\end{itemize}
Additional details and results from our work, video demonstrations of robot tasks and  summaries of this research can be found on the \href{https://arkadeepnc.github.io/projects/active_workspace/index.html}{project web page}.

%-------------------------------------------------------------------------
\section{Related Work}\label{sc:related_work}
% \section{Related Work and Motivation}\label{sc:related_work}
% In this section we review the body of literature relevant current work
\textbf{Shape from shading}, introduced by Horn\cite{Horn1970} and Woodham \cite{Woodham1980}, is a classical problem in computer vision which involves obtaining the shape and surface reflectance of an object by varying either the viewing direction, or the scene illumination or both. Several versions of this problem have been proposed where the researchers have recovered the shape, reflectance and shading of the object from learned priors~\cite{Barron2014}, proposed accurate methods for recovering object shapes as a collection of algebraic surfaces~\cite{Xiong2014}, and more recently have demonstrated highly accurate frameworks for recovering object shape, reflectance and geometry by refining multi-view RGB-D data captured by commercial sensors~\cite{Luan2021}.

Although a large portion of recent work \cite{Mildenhall2021,Boss2021,Srinivasan2021,Fridovich2022} advocates  learning implicit representations of objects from multiple views and then using them for robotics tasks~\cite{Yen2022nerf, Yen2021inerf},  there has been increasing interest in research on inferring object geometry from its interactions with controlled directional illumination. Recent work has demonstrated the capture of object surface normals and reflectances using multi-view photometric stereo ~\cite{Yang2022ps}, explicitly recovering object shape and reflectance from images taken with cameras equipped with flashes~\cite{Luan2021,Cheng2023wildlight} and have demonstrated cameras paired with an ensemble of projectors and flashes to capture scene properties~\cite{Schmitt2020, Higo2009}. Other notable efforts include learning implicit scene representations from multiple directionally illuminated images~\cite{Yang2022s3}, estimating object geometry from shadows cast by the object under directional illumination~\cite{Tiwary2022}, and reconstructing surface depth and normals from images under directional illumination~\cite{Karnieli2022}. Although a large portion of these works demonstrate impressive accuracy for a selected set of objects, it is unclear how appropriate any of these are as a perception system for manipulation tasks. \cite{Johnson2011} were the first to demonstrate the use of photometric stereo as a metrically accurate sensor and smaller versions of the setup~\cite{GelSight, GelSightMini} have been shown to be useful for several manipulation tasks. Our work draws inspiration from \cite{Johnson2011} as we demonstrate the  applicability of object geometry capture for different robotic manipulation tasks using  techniques from classical computer vision.

\section{Methods} \label{sc:methods}
It is well known from the shape from shading literature~\cite{Horn1970, Barron2014} that the measured intensity through an imaging device is a function of three major quantities -- the shape and reflectance of the object and the illumination of the environment. In this work, we focus on recovering surface normals as a proxy for the object shape. We use controlled lighting in addition to ambient illumination. Further, we simplify the shape from shading problem by exclusively considering Lambertian objects -- painting the objects with matte white paint so that the reflectance is known (Lambertian). 

As shown in \cref{fig:workspace_schematic}, we illuminate the workspace with a low angle of incidence with approximately parallel light rays $\mathbf{l}$, emulating a light source at infinity. Objects are also indirectly illuminated by the ambient light and we model the illumination as a combination of a linear model~\cite{Barsky2003}:
\begin{align}\label{eq:lambertian_shading}
I_i^k = \frac{\rho}{\pi}\langle\mathbf{n}_k, \mathbf{l}_i \rangle ~ ~ \forall i \in 1...,6
\end{align}
and a quadratic model $\mathbf{M}$ using second order spherical harmonic functions~\cite{Ramamoorthi2001,Johnson2011}:
\begin{align}\label{eq:quadratic_shading}
    I_i^k = {{\mathbf{n}}_k}^T\mathbf{M}_i{{\mathbf{n}}_k} ~ ~ \forall i \in 1...,6
\end{align}
At inference time, given measured pixel intensities $I_i^k$ and albedo $\rho$, and per-channel linear and quadratic illumination models $\mathbf{l}_i,~\mathbf{M}_i$, we obtain the surface normals $\mathbf{n}_k$ by inverting the models sequentially for $L_{1:6}$ in \cref{fig:workspace_schematic}, while accounting for shadows on the object. Optionally, we spatially integrate the surface normals to get a depth map of the object's surface. As our perception pipeline builds upon traditional techniques from photometric stereo, we defer a detailed discussion of our methods to the supplementary material. 
% ------------------------------------------------
\section{Experiments and Results}\label{sc:results}
In this section we describe our experiments on using our sensing system for common manipulation tasks. We start by quantifying the performance of our approach and then demonstrate the use of our approach in three sub-tasks related to manipulation -- general purpose object picking, estimating object deformation with vision, and pose estimation. We focus the paper on the results of our experiments, deferring details of the methods to the supplementary material.
\subsection{Performance of our sensing approach}\label{sc:sensor_performance}
% In this section we discuss the performance of our approach in perception tasks related to manipulation. We 1) compare the performance of our approach to approaches using several commercial sensors and 2) demonstrate the accuracy of our sensor in measuring the surface normals and depth of known objects. 
% We classify the performance indices into: 1.) how do we stack up against commodity sensors generally used for manipulation tasks and 2.) how accurate is our perception system in measuring object normals and inferring surface depth from the normals. 
\begin{figure*}
\centering
    \begin{subfigure}[b]{0.90\textwidth}
        \centering
        \includegraphics[width=\textwidth]{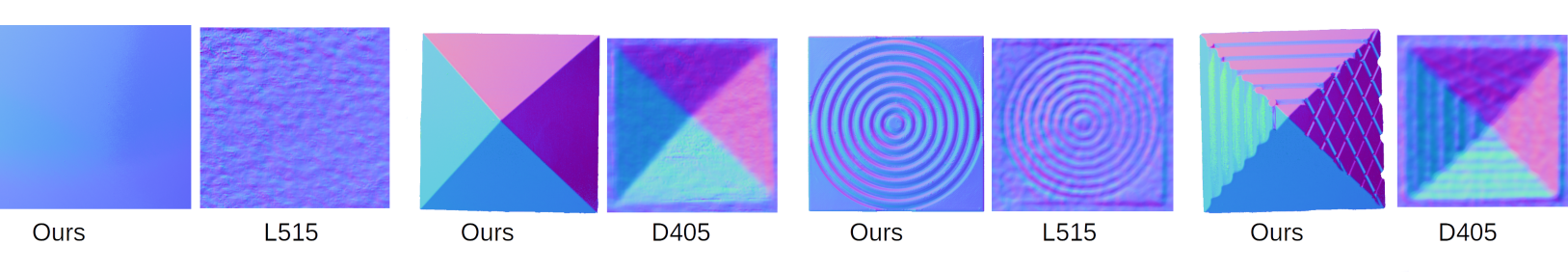}
        % \squeezeup
        \caption{}
    \label{fig:depth_sensor_comparisons}
    \end{subfigure}
    \begin{subfigure}[b]{0.90\textwidth}
        \centering
        \includegraphics[width=\textwidth]{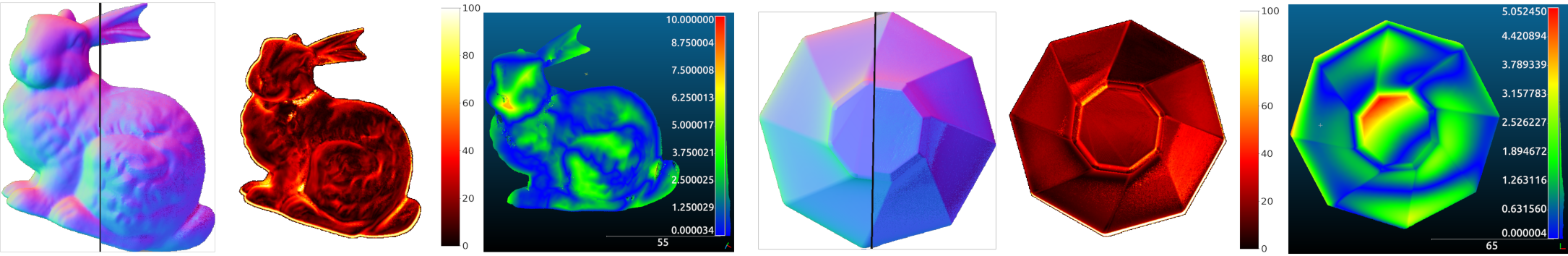}
        \caption{}
    \label{fig:qualitative_surface_comparison}
    \end{subfigure}
    \caption{\textbf{Our approach compared to commercial depth sensors}. In \cref{fig:depth_sensor_comparisons} we compare the quality of normals computed by our sensor with the processed data from the best performing commercial depth sensor (\cref{tab:sensor_meas_accuracy}). \Cref{fig:qualitative_surface_comparison} shows the normals of the bunny and octagon II from \cref{tab:sensor_meas_accuracy} as sets of 3 images each. In each set from left to right, we overlay the normals calculated by our approach (right of black line) on the ground truth normals (left of black line), the per pixel deviations of the normals from ground truth (in degrees) and the deviations of the measured surface from the ground truth mesh in mm.}
\label{comparison_master}
\end{figure*}
\subsubsection{Quality of normals versus true depth sensors}\label{sc:sensor_benchmarks}
% In this section we present a comparison between object normals captured by our setup versus normals synthesized from the depth images captured using commercial depth cameras. 
The commercial depth sensors widely used for robot manipulation tasks are fundamentally different from our sensor because the primary measurement in those sensors is the depth of the visible surface from the sensor calculated either through stereo matching (first 3 rows of \cref{tab:sensor_comparisons}), or through time-of-flight (rows 4, 5 of \cref{tab:sensor_comparisons}). For these depth sensors, we calculate the normals as spatial derivatives of the captured depth. Our sensing approach infers object surface normals from shaded images and optionally calculates a depth map that best explains the observed normals when the whole object is visible from the cameras' viewpoint. We also note that the commercial depth sensors we used (except for the D405) are designed to operate in room scale environments and are not necessarily suited for measuring surfaces of the objects we used. 
% A comparison of the depth maps across all the sensors, therefore, is not a common ground. 
To focus on the quality of the estimated surface normals we measure the statistical similarity of the normals measured across all the sensors on the same object patches. We image flat and textured surfaces at two orientations -- facing the camera's projection axis and at an inclination of 45$^\circ$ to the projection axis at the closest possible distance. 
% For each case, we capture the target object depths, compute normals as spatial gradients of depths, remove the mean of the angles subtended by the normals to the projection axis (z axis) and generate a histogram of all the residual angles the normal vector makes with the projection axis by quantizing the angles into 1$^\circ$ spaced bins in $[-45^\circ, 45^\circ]$. The removal of means prevents us from measuring the bias of the sensors, but we ensured that the sensors had low bias by verifying the the mean of the angles were around 0$^\circ$ and 45$^\circ$ for the flat and the inclined surfaces before removing the means. Comparing the distribution of the angles around the mean gives us a common ground to compare all the sensors by abstracting away the effects of widely varying camera field of views across all the cameras. We then compare the histogram for the particular case for a sensor with the ground truth histogram generated similarly and report the dissimilarity of the normalized histograms as a measure of accuracy of the sensors. 
We use the earth movers distance~\cite{Rubner2000,Ling2007EMD} to compare the normals captured by our approach and the commercial sensors to the ground truth. 

We also note that the stereo-based depth sensors rely on visual textures which our objects lack. To address this, we follow the recommendations for imaging textureless surfaces from \cite{Keselman2017} by projecting visible (D405) or infrared (D435, D455) patterns on the objects as applicable. Additionally, to reduce the noise in the measurements, for the stereo sensors (rows 1 through 3 in \cref{tab:sensor_comparisons}), we calculate the measured depth as a trimmed mean (we remove 10\% of smallest and largest outliers) of all the depths at each pixel for 50 consecutive frames (acquired over 1.5 - 2 seconds). For the time of flight sensors (rows 4 and 5 in \cref{tab:sensor_comparisons}) we use a $3\times3$ pixel window median filter to smooth the captured depths at each pixel after temporally filtering the data using trimmed means. We present our quantitative results in \cref{tab:sensor_comparisons} and our qualitative results in \cref{fig:depth_sensor_comparisons}. We outperform all the commercial sensors in imaging surfaces as normals in each category, often by significant margins. Our approach is much better at detecting that a surface is actually smooth and flat, and in measuring local surface orientation. Unsurprisingly, we also note that the D405 and L515 sensors outperform the other sensors because they were designed for (or support) close range imaging which is relevant to the task we tested the sensors on.
\begin{table}
    \centering
    \begin{footnotesize}
    \begin{tabular}{@{}l|cccc@{}}
    \toprule
         Sensor & Flat $0^\circ$ & Flat $45^\circ$ & Texture $0^\circ$ & Texture $45^\circ$  \\ \midrule
          \href{https://www.intelrealsense.com/depth-camera-d455/}{D455} & 0.12 & 0.12 &  0.27 & 0.12   \\
          % {D455} & 0.1210 & 0.1203 &  0.2692 & 0.1196   \\
           \href{https://www.intelrealsense.com/depth-camera-d405/}{D405} & 0.09 & 0.04 & 0.21 &  0.15 \\
           % {D405} & 0.0903 & 0.0386 & 0.2077 &  0.1471 \\
          \href{https://www.intelrealsense.com/depth-camera-d435/}{D435} & 0.09  & 0.23 & 0.22 &  0.22 \\
          % {D435} & 0.0908 & 0.2281 & 0.2240 &  0.2231 \\
          \href{https://www.intelrealsense.com/lidar-camera-l515/}{L515} & 0.06 & 0.05  & 0.20 & 0.15 \\
          % {L515} & 0.0556 & 0.0462  & 0.1970 & 0.1495 \\
          \href{https://azure.microsoft.com/en-us/services/kinect-dk/#overview}{Kinect II} & 0.24 & 0.14 & 0.19 & 0.16  \\
          % {Kinect II} & 0.2372 & 0.1396 & 0.1934  & 0.1641 \\
           % Ours & \textbf{0.0151} & \textbf{0.0084} & \textbf{0.1883} & \textbf{0.0962} \\
           \midrule
          Ours & \textbf{0.02} & \textbf{0.01} & \textbf{0.18} & \textbf{0.10} \\
         \bottomrule
    \end{tabular}
    \end{footnotesize}
    \caption{\textbf{Comparison of our approach with commercial depth sensors} (lower is better). The metric value indicates the \textit{dissimilarity of the measured normals with ground truth} which in the case of flat surfaces is related to the standard deviation of the angles of the normals with respect to the mean. Representative normal maps corresponding to the next best performing sensor have been provided in \cref{fig:depth_sensor_comparisons} along with the data captured by our sensor in those categories.}
    \label{tab:sensor_comparisons}
\end{table}

\subsubsection{Accuracy of our approach}\label{sc:accuracy_test}
\begin{table}[tb]
\centering
\begin{footnotesize}
\begin{tabular}{@{}l|cc|cc@{}}
\toprule
  & \multicolumn{2}{c}{Normals  ($\triangle^\circ$)} & \multicolumn{2}{c}{Surface ($\triangle$mm)} \\ \midrule
 Object name & $\mu (\sigma)$ & $<20^\circ$ (\%) & $\mu (\sigma)$ & max \\ \midrule
Pyramid & 13.31 (12.89) & 93.06 & 1.31 (0.93) & 4.17 \\
Star & 18.85 (14.04) & 86.73 & 0.80 (0.60) & 3.65 \\
Bent Cyl. & 18.81 (15.08) & 86.19 & 0.50 (0.36) & 2.03 \\
Octagon I & 16.23 (12.53) & 83.45 & 0.87 (0.85) & 4.50 \\
Octagon II & 17.17 (12.09) & 82.52 & 1.20 (0.90) & 5.05 \\
Spot & 17.38 (10.42) & 83.55 & 0.54 (0.48) & 2.82 \\
Bas-relief & 18.24 (10.01) & 76.24 & 0.87 (0.94) & 4.26 \\
Bunny & 20.07 (14.80) & 75.88 & 1.60 (1.27) & 8.12 \\
Text. Pyramid & 19.19 (17.52) & 61.75 & 1.74 (1.31) & 7.30 \\
Happy Budd. & 29.19 (20.52) & 54.29 & 1.54 (1.22) & 7.66 \\ \bottomrule
\end{tabular}
\end{footnotesize}
 \caption{\textbf{Quantitative measurements of our approach in estimating object shape.} The measurements in the first set of columns are the deviations of measured normals from ground truth normals in degrees. The measurements in the second set of columns are the deviation of the reconstructed surface from the ground truth mesh in millimeters. Please visit the project website for a qualitative visualization of the results.}
\label{tab:sensor_meas_accuracy}
\end{table}
% In this section we quantify the accuracy of our sensor in measuring object surface normals and object surface reconstruction. 
To quantify the accuracy of our sensor, we 3D printed a set of objects of footprint less than 12cm$^2$ using a standard 3D printer. % Prusa MK3S 3D printer (we used Prusa PLA filament and a layer height of 0.15mm) and spray painted the objects matte white. 
The objects were imaged with our setup in \cref{fig:opening_figure}. 
The surface normal quality measurement is performed by finding the object pose that aligns the measured surface normals to the ground truth surface normals generated with a renderer imaging the ground truth mesh. We used photometric stereo to calculate the normals and the method described in \cref{sc:localization} to align the captured object to the ground truth mesh model. We then calculated the angle between the measured normal and the rendered ground truth normals at every pixel and reported the statistics of the angles as a quantitative measure of the quality of the normals estimated by our approach. Our pose estimation pipeline (\cref{sc:model_based_pose_estimation}) aligned the simulated and the real data up to a maximum error of 5 pixels, making the per pixel error calculation meaningful. In \cref{tab:sensor_meas_accuracy} under the label `normal quality' we report the mean and the standard deviation of the per pixel angle error in degrees. We also report the percentage of pixels with angle error less than $20^\circ$. This metric was influenced by our observation during vacuum picking experiments (see \cref{sc:pickup_experiments}) that with the choice of a more compliant gripper, our gripping pipeline was tolerant of surface normal errors up to $20^\circ$. Notwithstanding the errors introduced due to warps on some 3D printed models (see e.g. high error pixels around the edges of insets 2 and 5 in \cref{fig:qualitative_surface_comparison}), the quality of our normal and depth estimates are similar to classical methods~\cite{Barron2014,Xiong2014} and are marginally worse than recent deep learning based methods~\cite{Karnieli2022, Yang2022ps, Yang2022s3}. 
\newline \indent To measure the surface quality, we generated a surface depth map by spatially integrating the normals of the scene calculated using photometric stereo (more details in the supplementary materials). We then registered the integrated depth map to the ground truth mesh using point-to-plane ICP~\cite{Rusinkiewicz2001} and calculated the Hausdorff distance~\cite{Cignoni1998METRO} between the recovered depth map and the ground truth mesh of the object. The mean, standard deviation and the maximum point-wise distance of the recovered depth map from the ground truth meshes are reported under the surface quality column of \cref{tab:sensor_meas_accuracy}. For both the experiments, we observe that the objects with large planar faces or smoothly varying curvatures worked the best while, objects with undercut surfaces performed poorly -- especially the happy Buddha object which had several undercut faces (see \cref{fig:model_based_pose_est}). We present a qualitative result in \cref{fig:qualitative_surface_comparison}. More qualitative results can be viewed on our project website. 

\subsection{Controlled illumination for pickup tasks}\label{sc:pickup_experiments}
To perform immobilizing grasps with a single suction cup vacuum gripper, we need to identify a portion of the object that is large enough for the suction cup to fit and flat enough for the suction cup to be effective. Additionally, we have to localize the point of grasp with respect to the gripper and identify the local surface orientation so that the gripper can approach along the surface normal with the face of the suction cup perpendicular to the surface to best execute the grasp. We have described pipelines to identify surface normals, demonstrated our system's performance in measuring surface discontinuities, and we have a stereo system ($C_1$ and $C_2$ in \cref{fig:opening_figure}) for triangulating a world point in the robot's frame. In this section, we describe our pipeline to detect arbitrarily oriented flat objects that can be grasped with a suction cup gripper of a given size. We also demonstrate how low angle of incidence directional illumination can help identify and pick up thin flat objects when segmentation is difficult using color or depth contrasts with conventional sensors. 
\begin{figure*}
\centering
\begin{subfigure}[b]{0.22\textwidth}
\centering
\includegraphics[width=0.85\textwidth]{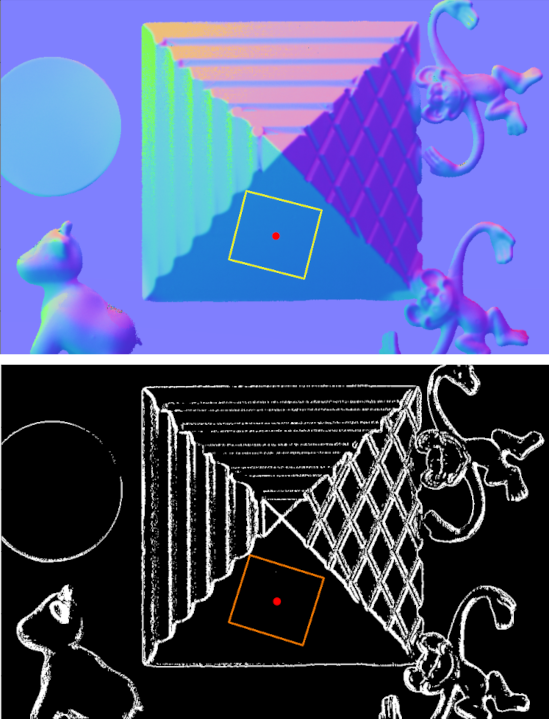}
\caption{}
\label{fig:12_mm_grasp_selection_45_deg}
\end{subfigure}
\begin{subfigure}[b]{0.22\textwidth}
\centering
\includegraphics[width=0.96\textwidth]{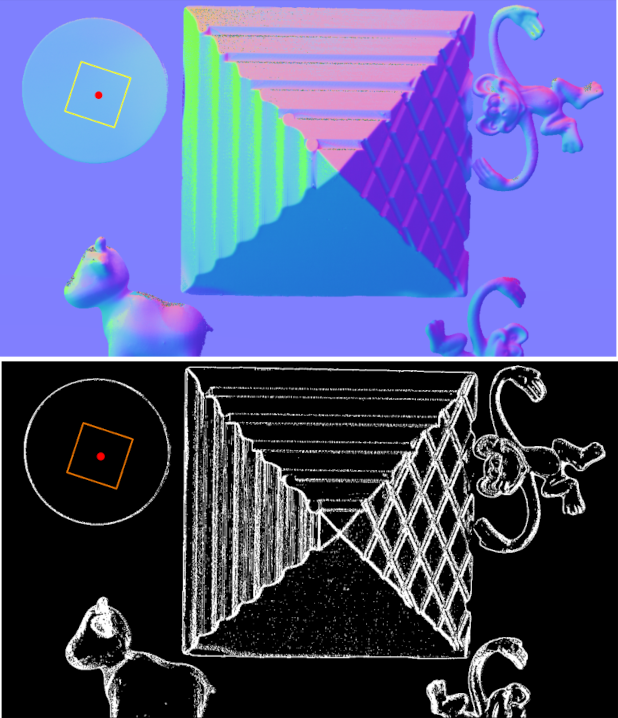}
\caption{}
\label{fig:12_mm_grasp_selection_flat}
\end{subfigure}
\begin{subfigure}[b]{0.35\textwidth}
\centering
\includegraphics[width=1.05\textwidth]{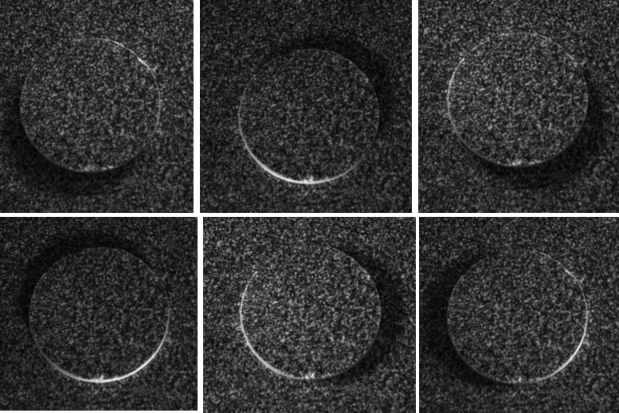}
\caption{}
\label{fig:dir_illum_flat_disc}
\end{subfigure} 
\begin{subfigure}[b]{0.16\textwidth}
\centering
\includegraphics[width=0.8\textwidth]{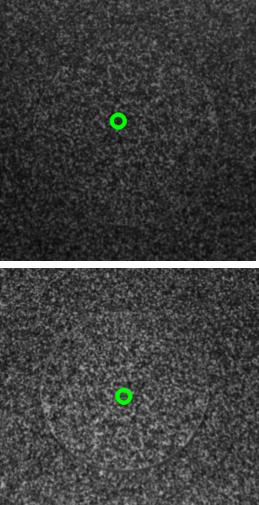} 
\caption{}
\label{fig:top_view_detected_disc}
\end{subfigure}
\caption{\textbf{Our pipeline for grasping objects in a robot's workspace.} \Cref{fig:12_mm_grasp_selection_45_deg,fig:12_mm_grasp_selection_flat} respectively show two distinct grasps selected by the camera $C_1$ along $X$ and $Y$ directions respectively. On the top rows of \cref{fig:12_mm_grasp_selection_45_deg,fig:12_mm_grasp_selection_flat} we show the detected grasps projected on the scene normals and, on the bottom row, we show the detected grasps projected onto the occlusion edge pixels of the scene. We note that our selected grasps do not have occlusion edges. \Cref{fig:dir_illum_flat_disc,fig:top_view_detected_disc} shows our pipeline for picking up thin objects in absence of obvious depth or texture segmentation cues. \Cref{fig:dir_illum_flat_disc} shows the images of a plastic disc (50mm diam., 3.15mm thick) captured with $C_1$ placed vertically above the disc, using the directional lights $L_{1:6}$. \Cref{fig:top_view_detected_disc} shows our detection of the center of the discs overlaid on the image captured with cameras $C_1$ (top) and $C_2$ (bottom) with $L_7$ and ambient light illuminating the scene.}
\label{fig:pickup_master}
\end{figure*}

% \indent \textbf{Picking up 3D Lambertian objects:} 
We divide the problem of picking up 3D objects into two major steps -- identifying the largest geometrically flat patch in the workspace across two views and executing the robot motion to grasp the identified face. To identify the largest corresponding flat patch across the camera views, we modified the well-known CAMShift algorithm\cite{Bradski1998} to filter out image patches not graspable with a vacuum gripper. 
% We describe our algorithm for identifying corresponding flat faces on objects in the robot's workspace across the two views $C_1$ and $C_2$ (see \cref{fig:workspace_schematic}) in Appendix \ref{sc:pickup_details}.

With geometrically corresponding grasping locations identified across two views, we use camera poses and intrinsics of the two views $C_1$ and $C_2$ to triangulate the position of the gripper in the robot's coordinate frame. For the orientation of the gripper -- we calculate the normal at the grasp location by averaging the normals at the identified grasp location in the two views. We then generate and execute a minimum jerk trajectory~\cite{Flash1985} which moves the robot to make contact with the selected object, picks the object up and, places it in a bin in the robot's workspace. 

\textbf{For picking up thin objects without color contrast} following Raskar et al. \cite{Raskar2004}, we observe that image brightness variations due to occlusion are more prominent than brightness differences due to color texture. We use this intuition to identify thin and flat objects from the background. For this experiment we pick a plastic disc of 50mm diameter, and 3.15mm thickness with a random pattern (pixel values sampled uniformly between 0 to 255) printed on the surface of the disc from a flat plane with the same visual texture. As is obvious from \cref{fig:dir_illum_flat_disc,fig:top_view_detected_disc}, there isn't enough intensity or color texture to segment the object from the background. However, looking along the light direction (see \cref{fig:dir_illum_flat_disc}), we note that the shadows due to occlusion are more prominent than the visual textures on the object and the background -- as imaged in \cref{fig:dir_illum_flat_disc,fig:top_view_detected_disc}. 

This shadow cue was sufficient for our edge detection procedure (see supplementary materials for details) to identify occlusion edge pixels in the scene. We then estimated the center of the disc by fitting a minimum enclosing circle~\cite{OpenCV2021} to the occlusion edge pixels across the two views $C_1$ and $C_2$. We visualize this in \cref{fig:top_view_detected_disc} -- the centers of the circles have been projected onto the images captured by the two views with the ambient and overhead light ($L_7$ in \cref{fig:workspace_schematic}). 
% However, in the presence of surface textures, we could not reliably detect normals on the surface of the object, so we assume that the disc lies flat on the workspace. 
To pick up the disc, we approached the center of the disc along the surface normal of the tabletop by triangulating the object from the two views in \cref{fig:top_view_detected_disc}. None of the commercial depth sensors we used (see \cref{tab:sensor_comparisons}) detected the disc reliably.

We performed 35 grasp experiments with 3D objects and ten experiments with flat textured objects (5 with the disc, and 5 with an irregular octagon inscribed by a 50mm diameter circle and of 3.15mm thickness). Movies of our experiments can be viewed on the project website. Across all the experiments we failed four times while picking up the textured pyramid (\cref{fig:12_mm_grasp_selection_45_deg}) due to the gripper failing to attach or colliding with the object due to triangulation errors. 

\subsection{Measuring deformation}\label{sc:card_bending}
\begin{figure*}
\centering
\begin{subfigure}[b]{0.20\textwidth}
\centering
\includegraphics[width=0.65\textwidth]{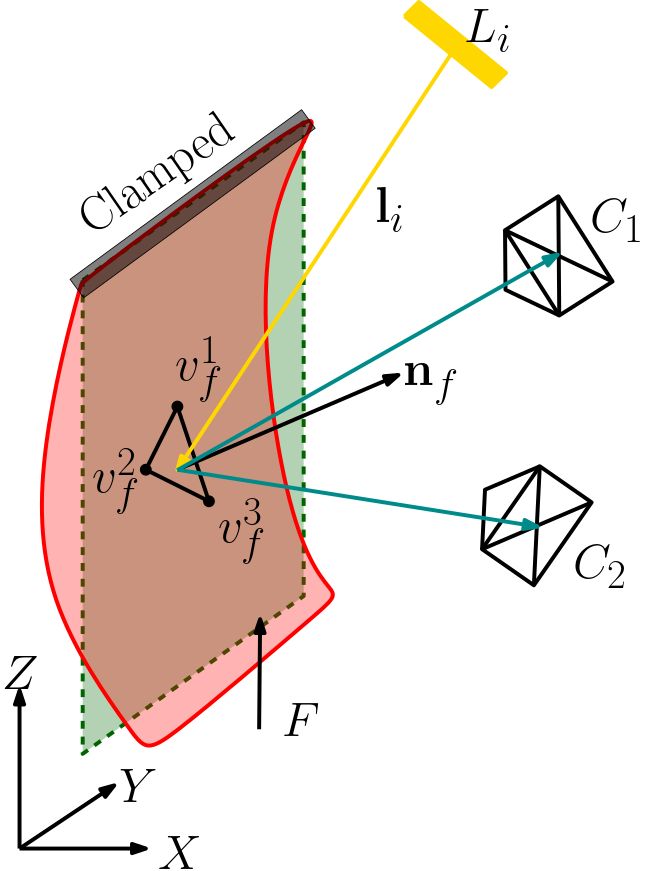}
\caption{}
\label{fig:card_bending_schematic}
\end{subfigure}
\begin{subfigure}[b]{0.40\textwidth}
\centering
\includegraphics[width=\textwidth]{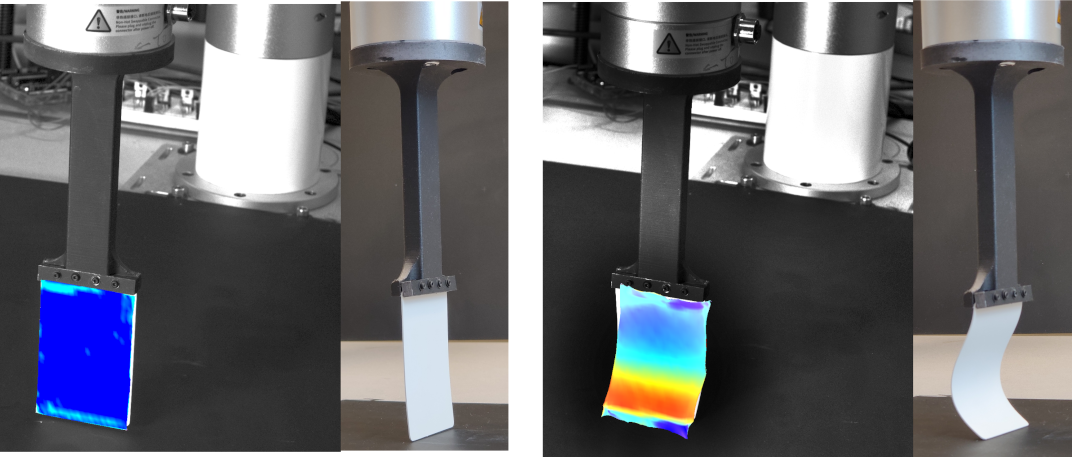}
\caption{}
\label{fig:card_bending_overlays}
\end{subfigure}
\begin{subfigure}[b]{0.32\textwidth}
\centering
\includegraphics[width=\textwidth]{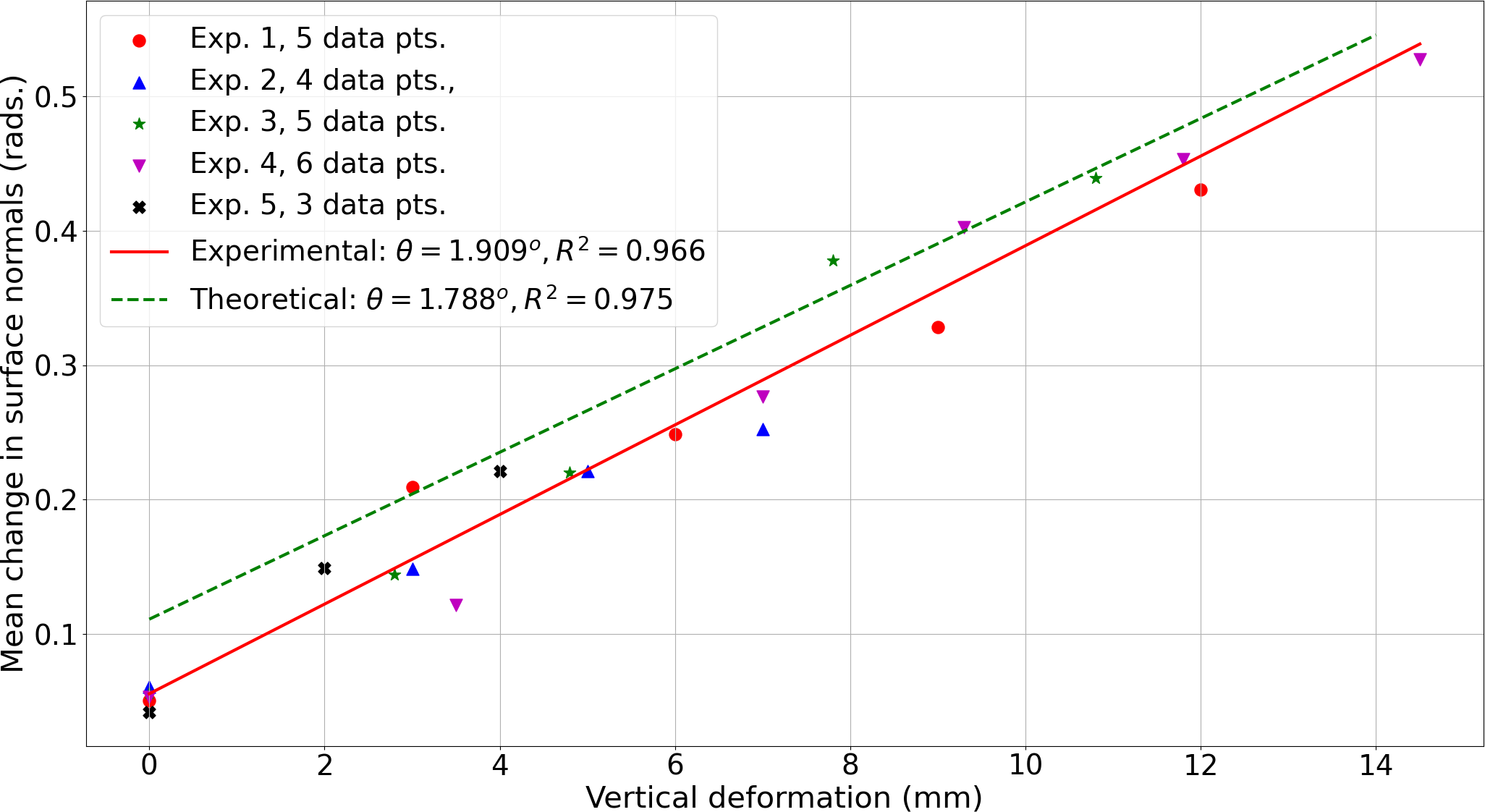}
\caption{}
\label{fig:card_bending_plots}
\end{subfigure}
\caption{\textbf{Our results of measuring the deformation of known objects.} \Cref{fig:card_bending_schematic} shows a schematic diagram of our procedure, \cref{fig:card_bending_overlays} shows the estimated shape of the card overlaid on a camera image. The insets of the images in \cref{fig:card_bending_overlays} provide an external view of the state of the cards. \Cref{fig:card_bending_plots} summarizes the quantitative results of result of our experiment to show we were able to estimate the physical process behind the deformation of the card.}
\label{fig:bending_master}
\end{figure*}
Our approach can be used to perceive the deformation of objects by tracking the change of local normals on the object's surface. In this section, we describe our method for measuring the buckling deformation of a 0.76 mm thick rectangular PVC card of size 86$\times$54mm.  We measure the deformation in a typical analysis-by-synthesis fashion -- we explain the change in the surface normals of the deformed card  by calculating the deformation of the card geometry. %We describe our pipeline in detail in Appendix \ref{sc:card_bending_details}.
% We start by describing our pipeline for capturing the deformation of the card and then describe our experiment that validates our claim that our pipeline does in fact capture the physical process of elastic deformation under certain mechanical assumptions. Although we demonstrate our pipeline for measuring deformation on a simple experiment, we ensure that the fundamental construction of the pipeline is independent of object geometry -- thus enabling generalization of our pipeline possible with minimal changes. 

\Cref{fig:card_bending_overlays} shows a qualitative result of our pipeline -- the first image shows the reconstructed shape before the onset of buckling and the second image shows the reconstructed mesh overlaid on the deformed object. In both cases, we show the local change in surface normals of the reconstructed shape as the color of the mesh. An external view of the objects is also provided as the inset to the images. During our experiments we observed that skewed viewing directions of the cameras $C_1$ or $C_2$, e.g. the inset views in \cref{fig:card_bending_overlays}, significantly reduced the performance of our pipeline due to the decrease in the effective number of pixels imaging the object across the two views. This led us to generally use $C_1$ and $C_2$ to capture frontal views. More qualitative results of our experiments can be viewed on the project website. Detailed explanation of our pipeline can be found in the supplementary materials. 

To verify that we indeed captured the physical process behind the card buckling under vertical loads, we compared our predictions with theoretical values calculated using solid mechanics.  The theoretical data showed a linear trend in the mean change in surface normals and the vertical deformation of the card. A line with a slope of 1.78$^\circ$ and offset of 0.114 fits the theoretical data with a $r^2$ score of 0.975. We repeated the  experiment of deforming the card five times with five different PVC cards and different camera positions ($C_1$ and $C_2$ in \cref{fig:card_bending_schematic}) and obtained 23 data points as shown in \cref{fig:card_bending_plots} with five different markers corresponding to the experiments. The experimental data also showed a linear trend -- a line with slope of 1.9$^\circ$, offset of 0.055 fit the data with a $r^2$ score of 0.966. Disregarding the difference in the offset due to the baseline noise in our measurements and the slight violation of a geometric boundary condition due to slipping of the card at the base, from \cref{fig:card_bending_plots} we can conclude that our pipeline is  repeatable and can capture the physical process of the buckling deformation of the card due to vertical loads.

% to perform the following experiment. We note that any vertical section along the longest dimension of the card can be modeled as a fixed-pinned beam with the following equation
% \begin{align}\label{eq:card_model}
% y(x) = \delta_{max}\left(1-\cos\dfrac{\pi x}{0.7(L_{max} - p)}\right)
% \end{align}
% where $\delta_{max}$ denotes the maximum deformation at any portion of the card, $L_{max}$ is the maximum length of the card (86mm in our case) and, $p$ is the deformation of the card along the $Z$ direction in \cref{fig:card_bending_schematic}. As the card is pushed down, the card buckles and the value of $p$ increases, but $L_{max}$ remains constant at 86mm. We solve for $\delta_{max}$ for 15 values of $p$ ranging from 0mm (no deformation) to 14mm. We uniformly sample each of the 15 curves generated by substituting the values of $p$ and $\delta_{max}$ into \cref{eq:card_model} at 35 points along the theoretical beam's length. We chose the 35 points to align with the 35 vertices along the 86mm edge of $\mathcal{M}$. We then calculated the mean of the change in normals at each sampled point between the undeformed and the deformed curve. Next, we calculated the surface normals of the deformed card using the pipeline discussed above and compared the mean change in surface normal orientation from the undeformed configuration of the card in \cref{fig:card_bending_plots}.\newline
\begin{table}
    \centering
    \begin{footnotesize}
    \begin{tabular}{@{}l|cccc@{}} 
    \toprule
         Method & Scale & Error $\mu (\sigma)$ & Max & Sample size  \\ \midrule
          VIRDO~\cite{Wi2022virdo} & $89\times56$ & 1.12 (- - -) & -- & 5.6k \\ 
          VIRDO++~\cite{Wi2022virdo++} & $89\times56$ & 1.04 (0.35) & -- & 5.6k \\ \midrule
         Ours & $86\times54$ & 0.78 (0.52) & 4.57 & 100k \\ 
    \bottomrule
    \end{tabular}
    \end{footnotesize}
    \caption{\textbf{Comparison of our method of measuring deformation} (chamfer distance) against baselines~\cite{Wi2022virdo,Wi2022virdo++}. Our experiment (single view) was carried out in simulation because we do not have access to a sensor accurate enough to measure ground truth deformations, and we are comparing the performance of Wi and colleagues' results during inference on simulated data with a single camera view. Our experiment captures the deformed card at a simulated endpoint deflection of 12 mm corresponding to a mean change in surface normals of 0.41 radians. 
    % ($^*$)We assume Wi et al. use typical spatulas in their work. 
    All length measurements are in mm. and all the experiments assume full visibility of the object.}
    \label{tab:compare_virdo}
\end{table}

 We note that in this work we exclusively use vision. We do not factor in the force that is causing the deformation in the object. In related work by Wi et al. \cite{Wi2022virdo,Wi2022virdo++} the authors combine visual measurements with forces measured by a force torque sensor coupled to the deforming object to infer its deformation. For objects of similar scale, we compare our performance with the works of Wi and colleagues in \cref{tab:compare_virdo}. Although we perform better in reconstructing shapes, we require nearly full visibility of the object in both views which is not a limitation for Wi et al. in \cite{Wi2022virdo++}. 
 
 We also note that our ``analysis-by-synthesis" procedure  works better than integrating the captured normal map (see supplementary for details) to reconstruct the surface of the card. This is due to the strong view dependence of the shape reconstruction, which prevents us from trivially incorporating multiple views to reconstruct the bent object. Using two camera views and an initial mesh of the object to predict the shape of the deforming card makes our pipeline more robust to shadows and slight occlusions in camera views.

\subsection{Controlled illumination for localization}\label{sc:localization}
\begin{figure*}
\centering
\begin{subfigure}[b]{0.48\textwidth}
\centering
\includegraphics[width=\textwidth]{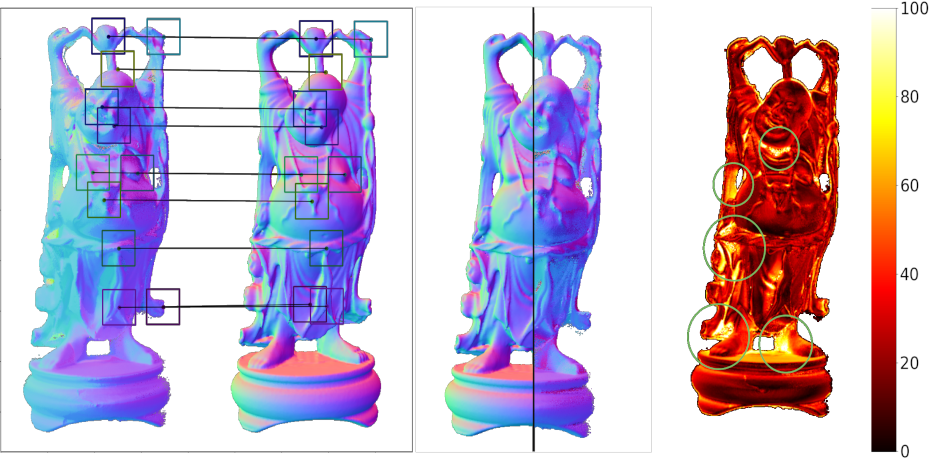}
\caption{}
\label{fig:model_based_pose_est}
\end{subfigure}
\begin{subfigure}[b]{0.48\textwidth}
\centering
\includegraphics[width=0.8\textwidth]{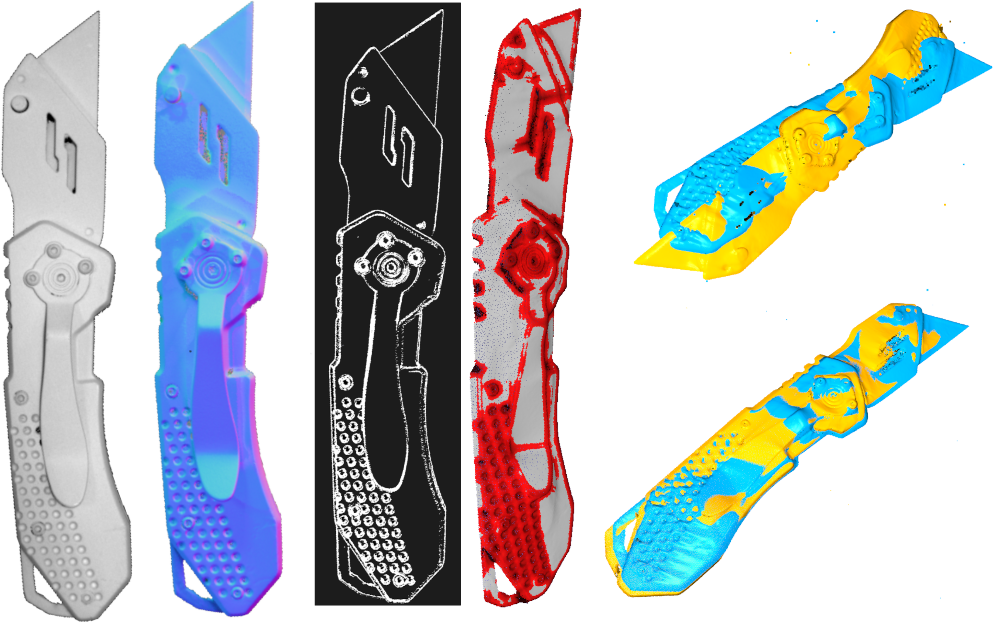}
\caption{}
\label{fig:model_free_pose_est}
\end{subfigure}
\caption{\textbf{Estimating object pose in the robot workspace.} \Cref{fig:model_based_pose_est} shows our steps in estimating object poses when a 3D model of the object is available apriori. From left to right in \cref{fig:model_based_pose_est} we calculate the pixel-wise correspondences between captured data and the available geometry, overlay the captured surface normals (right of black line) onto the ground truth object normals (left of black line) and show our pixel-wise pose estimation costs about an arbitrary scale. The green circles indicate areas with high local costs due to errors in estimating normals. \Cref{fig:model_free_pose_est} shows our pipeline for tracking object pose when a 3D model is not available a priori. From left to right, we show our captured data, calculated normals, calculated depth edges, calculated 3D representation of the surface with salient point-features overlaid on the points in red and, initial and final stages of our estimation of the object's pose.}
\label{fig:pickup_master}
\end{figure*}
% So far we have demonstrated our pipelines for capturing object normals, depth edges and upgrading object normals to metric depth maps of the object surfaces. 
% In this section, we describe our results for estimating object poses from data captured by our system.  
\subsubsection{Estimating poses of known objects}\label{sc:model_based_pose_estimation}
When an object model is available, we estimate its pose by aligning the measured surface normals, depth edges, and object silhouettes of the object with their simulated counterparts. We improve on commonly employed pipelines in the pose estimation literature ~\cite{Chaudhury2022,Liu2012,Imperoli2015} by generating good initial pose estimates through discovery of object patch correspondences between the observed and simulated surface normals (inset 1 of \cref{fig:model_based_pose_est}). We describe our method in detail in the supplementary material. 

We present a qualitative result in \cref{fig:model_based_pose_est} -- the second inset overlays the normal images with measured normals $\mathbf{N}_R$ on the left of the black line and rendered normals $\mathbf{N}_S$ on the right. In the third inset of \cref{fig:model_based_pose_est}, we visualize our per-pixel pose estimation costs overlaid on the observed image silhouette $\mathbf{M}_S$ on an arbitrary scale between 1 to 100. We note that the under-cut parts of the geometry, highlighted by green circles, have large local  costs, but we do not incur large costs due to silhouette misalignment or scaling as seen by the absence of high cost patches at the object edges or background. For all the objects tested, our multi-scale and multi-modal pose estimation pipeline reliably converges to the correct pose within a 5.5 pixel error (7 px/mm) even with high local errors in the measured data -- the model in \cref{fig:model_based_pose_est} has $\sim$50\% of pixels with erroneous normals ($>20^\circ$ deviation from ground truth).

We present quantitative measurements of our pose estimation approach for all the objects in \cref{tab:sensor_meas_accuracy} and \cref{tab:model_based_pose_est}. Our experiments were done with $C_1$ imaging a 227mm $\times$ 129mm area at a resolution of 7 pixels/mm. We placed each of the objects at a random position and orientation and used the pipeline described in this section to align the 3D model to the observed data. After the alignment, we overlaid the simulated and measured data and manually measured the pixel misalignment between the real and simulated data around the object edges. We repeated this experiment five times for each object and report the mean and standard deviations of the pixel misalignment in \cref{tab:model_based_pose_est}.
\begin{table}[tb]
    \centering
    \begin{footnotesize}
    \begin{tabular}{@{}l|cc@{}}
    \toprule
         Object & Bounding Box (mm) & Px. Errors $\mu (\sigma)$ \\
         \midrule
         Pyramid & $100 \times 100\times 55$ & 2.50 (1.70)\\
         Star & $110 \times 110\times 15$ & 3.62 (1.03) \\
         Bent Cylinder & $116 \times \:48 \times 13$ & 5.50 (2.12)  \\
         Octagon I & $100 \times 100\times 29$ & 3.35 (0.88)\\
         Octagon II & $100 \times 100\times 29$  & 2.85 (1.83) \\
         Spot & $\:52 \times \:15\times 53$ & 1.50 (0.23)\\
         Bas-relief & $\:83 \times 110\times \:7$ & 0.80 (0.20)\\
         Bunny & $\:93 \times \:94\times 43$  & 4.06 (1.19)\\
         Text Pyramid & $100 \times 100\times 55$ & 4.02 (0.96)\\
         Happy Buddha & $115 \times \:45 \times 37$ & 3.10 (0.69)\\
         \bottomrule
    \end{tabular}
    \end{footnotesize}
    \caption{\textbf{Performance of pose estimation for known objects.} All the experiments were done with a resolution of $\sim$7 pixels/mm. Qualitative results for the objects happy Buddha, bunny and octagon II can  be found in \cref{fig:model_based_pose_est} and the first and fourth insets of \cref{fig:qualitative_surface_comparison} respectively. More qualitative results can be found on the project website.}
    \label{tab:model_based_pose_est}
\end{table}

\subsubsection{Estimating the pose of unknown objects}\label{sc:model_free_pose_est}
If a 3D model of an object is not available, the pose estimation problem, defined classically, is ill-posed. In those cases, we can estimate the object's change in pose through rigid registration of the 3D representations of the same object as it moves. For our 3D representation, we choose the point cloud obtained by integrating the surface normal maps, factoring in the camera intrinsics (see supplementary for details). As noted before, this representation is not metrically correct for parts of the object that are not fully visible by the camera (e.g. full profile of the belt clip of the knife in \cref{fig:model_free_pose_est}), however, the relative surface depth changes and the overall scale of the object are captured accurately which lets us estimate the change in pose between two measurements of the same object. We pictorially describe our pipeline for tracking unknown objects in \cref{fig:model_free_pose_est}, which involves capturing the image of the object, measuring its surface normals and depth edges, generating a point cloud of the observed surface and registering two instances of the point clouds to obtain the change in pose. We describe our pipeline in detail in the supplementary material.

To evaluate our pipeline for estimating pose changes of unknown objects, we imaged four objects of different scales twice, while introducing a known pose perturbation between two measurements and recovered the pose perturbation using the method described in this section. We repeated this experiment six times for each of the objects in \cref{tab:model_free_pose_est} and report our pipeline's uncertainty in recovering the pose perturbations. We present our quantitative results in \cref{tab:model_free_pose_est} -- our approach has a tracking uncertainty of about $2^\circ$ in planar rotation and about 4 mm in translation.  
\begin{table}
    \centering
    \begin{footnotesize}
    \begin{tabular}{@{}l|cccc@{}}
    \toprule
         Object & Bounding Box (mm) & $\triangle X$(mm) & $\triangle Y$(mm) & $\triangle \theta (^\circ)$ \\
         \midrule
         Knife & $172\times 30\times 20$ & 1.02 & 1.66 & 0.36  \\
         Monkey & $\:46\times 64\times \:8$ & 1.67 & 0.97 & 0.22  \\
         Circuit & $\:13\times21\times 33$ & 3.28 & 2.40 & 1.28  \\
         IO shield & $120\times 39\times 21$ & 1.48 & 1.86 & 0.25  \\
         \bottomrule
    \end{tabular}
    \end{footnotesize}
    \caption{\textbf{Performance of our tracking pipeline for unknown objects.} All the experiments were done with a resolution of approximately 7 pixels/mm. Qualitative results for the knife is shown in \cref{fig:model_free_pose_est}, to view results of the other objects, please visit the project website. }
    \label{tab:model_free_pose_est}
\end{table}

Counter-intuitively, we also noted that generating a mesh of the object from the captured 3D representation and then using the mesh in the pipeline discussed in \cref{sc:model_based_pose_estimation} actually led to poorer pose estimation because the successive processing steps (normal calculation, integration, and meshing) reduced the quality of the mesh input. Since the 3D representation generated is strongly dependent on the viewpoint of the camera during the experiment meant that the 3D model was metrically incorrect for novel views. % other than the one used to generate it, thus making the resulting mesh unfit for the previous pipeline.

%-------------------------------------------------------------------------

\begin{figure}[thb]
    \centering
    \includegraphics[width=1\linewidth]{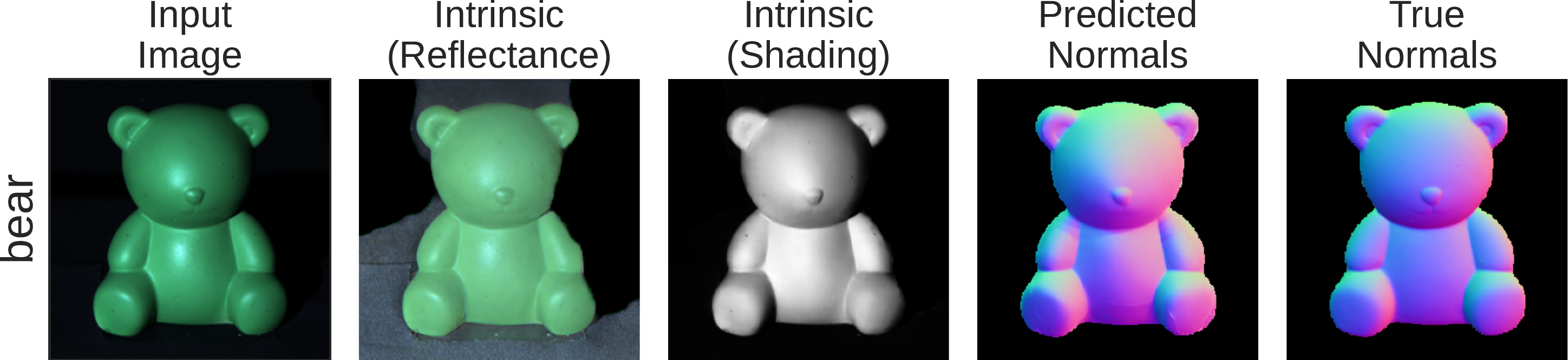}
     \includegraphics[width=1\linewidth]{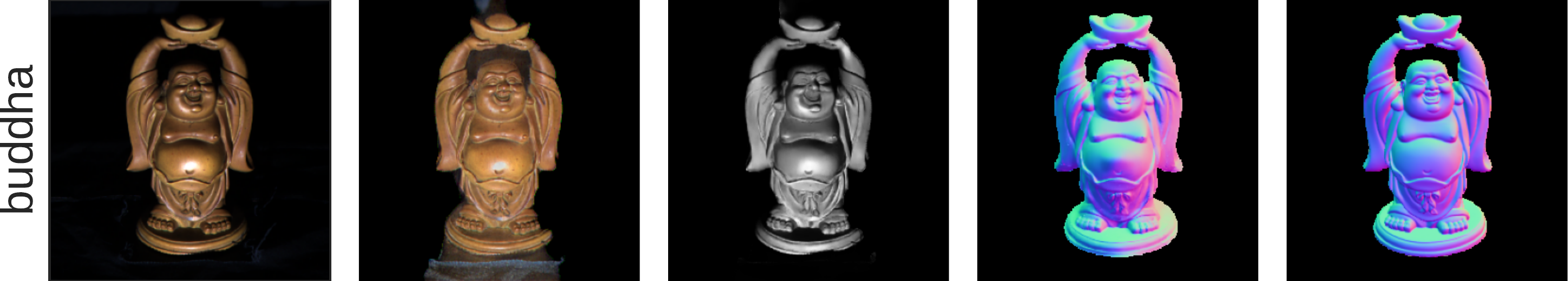}
    \includegraphics[width=1\linewidth]{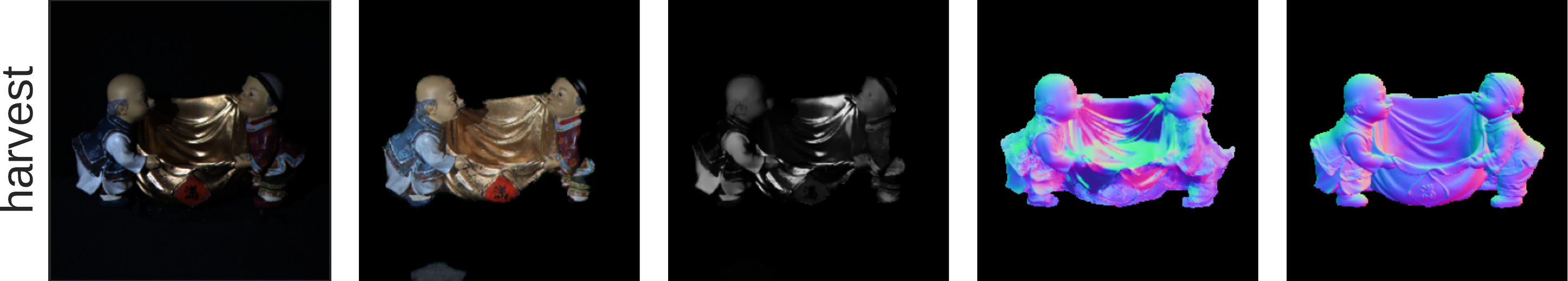}
   \caption{\textbf{Our photometric stereo on colored, non-Lambertian objects} using intrinsic image decomposition~\cite{Das2022PIE-NET} on a dataset~\cite{shi2019_uncalibratedphotometric}. \cref{sec:pienet}.}
   \label{fig:pienet}
\end{figure}
\subsection{Application to generic objects} \label{sec:pienet}
Our previous experiments focused on Lambertian objects with diffuse, white paint in calibrated environments. Extending these techniques to more general objects is possible using off-the-shelf methods for intrinsic image decomposition. Intrinsic image decomposition separates natural color images into a reflectance image and a shading image. These shading images can be used as input to any typical photometric stereo method, as they estimate the isolated impact of lighting across the surface. There are many approaches to intrinsic image decomposition, including classic priors\cite{Barron_2013_CVPR} and data-driven learning methods~\cite{Baslamisli_2018_CVPR,intrinsic_forysth2022}. In our experiments, we use a recent method called PIE-Net\cite{Das2022PIE-NET} to perform the intrinsic image decomposition. 

We show the results of this approach on three examples from the DiLiGenT\cite{shi2019_uncalibratedphotometric} dataset, which contains several objects that include not only color but also  general non-Lambertian materials. We show the results of a simple pipeline, using the pretrained PIE-Net model to obtain shading images. Photometric stereo is solved using the global, linear model that assumes distant point light sources and Lambertian shading. For each scene, DiLiGenT provides a single view of the scene with 95 different illumination conditions. We used 30-40 lighting directions for our reconstructions, each at PIE-Net's $256^2$ image resolution. 

Our results are shown in \cref{fig:pienet}. While this pipeline produced good results for some objects (bear, buddha), it struggled with the  glossy, dark, and self-occluding harvest object.   Nonetheless, considering that our pipeline used an off-the-shelf network and classic photometric stereo (without explicit handling of shadows or specularities), the results are promising. While this approach is far from the metric quality of our robotics setup in \cref{sc:accuracy_test,sc:model_free_pose_est,sc:model_based_pose_estimation}, we believe this quality of normals is potentially promising for vaccum gripper tasks such as those in \cref{sc:pickup_experiments} or the deformation estimates in \cref{sc:card_bending}. While we focused on the high end of precision in photometric stereo, this type of off-the-shelf pipeline enables generalization to generic objects, and produces potential utility in less demanding tasks. 
% Darkness is a common issue, as PIE-Net expects 8-bit input images, but DiLiGenT's images use all 16-bits of dynamic range.

%-------------------------------------------------------------------------

\section{Discussion and Future Work}\label{sc:discussion}
Having a Lambertian reflectance requirement for our objects is restrictive and can be a barrier for our methods to apply to many manipulation tasks -- future work can build on \cref{sec:pienet} to enroll general objects into our pipeline. 
% In \cref{sec:pienet} We envision combining a neural representation for reflectances with our physically based method for capturing shape to extend our method to general objects. 
During this work, we observed that although we do well in inferring the cumulative shape of the object we often have high local errors  -- see e.g. the second inset of \cref{fig:qualitative_surface_comparison}, and third inset of \cref{fig:model_based_pose_est}. We believe that these local errors are due to shadows that could not be resolved by our proposed method of inferring shape at a single pixel level with lights that are not co-incident with the camera. Further research is required on multiplexing the illumination sources to reduce the effect of shadows and to determine a better combination of light and camera locations. Lights collocated with cameras and on the robot workspace will possibly address some of the limitations of our work. Further research is also required for selecting alternative object representations. 
% The coverage of a pixel is dependent on the resolution of the camera and the relative pose of the camera and the object of interest, and we observed that the accuracy of our methods fall sharply when the image resolution was decreased. 
Current literature indicates that a triangle-based representation~\cite{Luan2021} or locally smooth patch-based representations~\cite{Xiong2014} may work but will need a plethora of hand-tuned regularizers, hyper-parameters and a significant amount of computational effort to converge to a meaningful representation. Volumetric representations~\cite{Mildenhall2021, Fridovich2022} have certain advantages over patch-based representations in the context of robotic manipulation. Integrating volumetric representations with a robot workspace scaled controlled illumination approach is future work.  
%Finally, the question of how much fidelity of measurement is needed for manipulation remains unanswered across the present literature. 
In this work, we achieve a higher fidelity of measurement than some commercial depth sensors by imposing priors on object reflectances and controlling illumination. %
%, but it is unclear if general-purpose robotic manipulation needs a high precision perception system for common manipulation tasks. 
A natural extension of the current work would be to augment the performance of a commercial depth sensor by using it in conjunction with our approach.

%--------------------------
\section{Conclusions}\label{sc:conclusion}
In this work we demonstrated the application of classical techniques from photometric stereo to robotic manipulation through a robot workspace scaled controlled illumination system. We showed that, by enforcing a reflectance prior on the objects, reasoning about observed object intensities conditioned on the direction of illumination can yield accurate surface normals and identify surface depth discontinuities with very little computation. We also showed that the normals captured by our approach are significantly better than the ones derived from measurements with commercial depth sensors and we also evaluated the accuracy of our approach in capturing surface depth and normals. With the surface representations generated using our approach, we demonstrated three common manipulation tasks -- picking up objects of arbitrary shape with a single point vacuum gripper, estimating bending deformation of a known object and estimating poses of Lambertian objects. 

\newpage
%%%%%%%%%%%%%%%%%%%%%%%%%%%%%%
% SUPPLEMENTARY 
%%%%%%%%%%%%%%%%%%%%%%%%%%%%%%

\title{Supplementary Material: Shape from Shading for Robotic Manipulation}
\author{}
\maketitle
%%%%%%%%% BEGIN

The paper website with a narrated summary of the research and interactive demonstrations can be accessed on the \href{https://arkadeepnc.github.io/projects/active_workspace/index.html}{project web page}.

\section{Methods}\label{sc:methods}
It is well known from the shape from shading literature (e.g. \cite{Horn1970, Barron2014}) that the measured intensity through an imaging device is a function of 3 major quantities -- the shape and reflectance of the object and the illumination of the environment. 
In this work, we focus on recovering surface normals as a proxy for the object shape. We use controlled lighting in addition to ambient illumination. Further, we simplify the shape from shading problem by exclusively considering Lambertian objects. 
% painting the objects with matte white paint so that their reflectance function is known (Lambertian). 
Future work will jointly estimate object normals and reflectance functions.
% In this work, we focus on recovering the shape of the  object by designing and measuring the illumination and fixing the reflectance of the objects' surfaces by forcing them to behave approximately as Lambertian materials by painting them with matte white paint. 
In this section, we discuss how we design and measure the illumination of the environment in \cref{sc:calib_light_table} and our method of capturing the shape of the object in the form of surface normals in \cref{sc:normal_recovery}. Se start with a detailed description of our  hardware setup.   

\subsection{Details of the hardware platform}\label{sc:hardware}
% \subsection{Hardware platform}

Our controlled illumination experimental setup (\cref{fig:workspace_schematic})\footnote{figure numbers have been carried over from the main document} consists of seven 25cm$\times$10cm light panels fabricated from commercially available LED strips\cite{waveform_lighting} fixed in the robot workspace as shown in \cref{fig:workspace_schematic}. Six light panels are placed around the robot's base to illuminate the robot's workspace with low angles of incidence. One panel is placed over the workspace to provide baseline illumination to calculate shadows (more details in \cref{sc:normal_recovery}). Each of the panels has a rated power of 45W, and is powered by a 500W switching mode power supply. The light panels are driven by a high-current switching transistor controlled with an Arduino Uno$^{\rm{TM}}$. For this work, the illumination of the room due to the ceiling lights, interreflections of all the lights around the ceiling and walls etc. are assumed to be constant. The Arduino relays the control commands from our algorithm running on a workstation to the light panels through a serial port. To capture images, we use two FLIR machine vision cameras ($C_1$ and $C_2$ in \cref{fig:workspace_schematic}) \cite{grasshopper3_usb3}. The camera $C_1$ is fitted with a 12mm focal length lens and camera $C_2$ is fitted with a 16mm focal length lens\cite{edmund_optics}. The cameras are configured to respond linearly to the amount of light captured by the lenses (gamma = 1) and output $1536\times 1536$ pixels 16 bit monochrome images. To position the cameras and perform manipulation tasks we use an xArm7 manipulator\cite{ufactory} and our vacuum gripper is manufactured from standard $1/4''$ vacuum fittings. Robot demonstrations are recorded with two HD webcams placed around the workspace. The data captured is processed on a Linux workstation with an Intel Corei9 processor, 64GB RAM, and an Nvidia RTX3090Ti graphics card with 25GB of vRAM. 

\begin{figure*}
\centering
\begin{subfigure}[b]{0.32\textwidth}
\centering
\includegraphics[width=\textwidth]{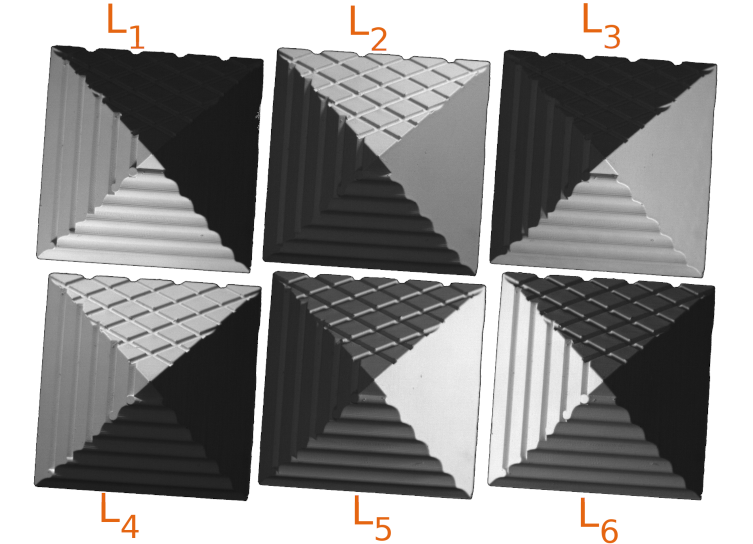}
\caption{}
\label{fig:light_channels}
\end{subfigure}
\begin{subfigure}[b]{0.17\textwidth}
\centering
\includegraphics[width=0.6\textwidth]{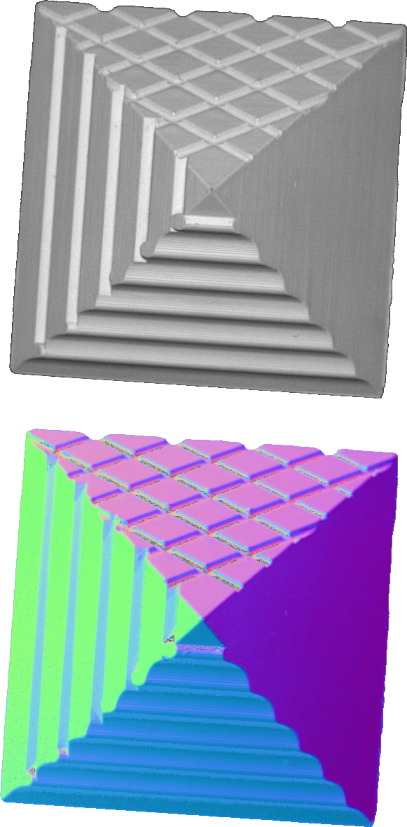}
\caption{}
\label{fig:top_channel_and_normal}
\end{subfigure}
\begin{subfigure}[b]{0.25\textwidth}
\centering
\includegraphics[width=\textwidth]{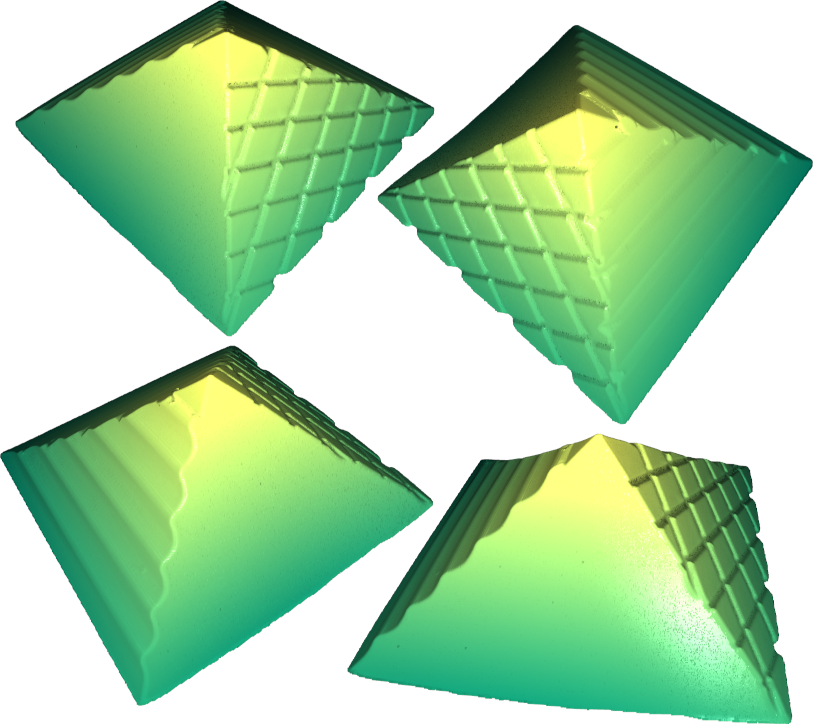}
\caption{}
\label{fig:pcd_textured_pyramid}
\end{subfigure}
\begin{subfigure}[b]{0.16\textwidth}
\centering
\includegraphics[width=0.8\textwidth]{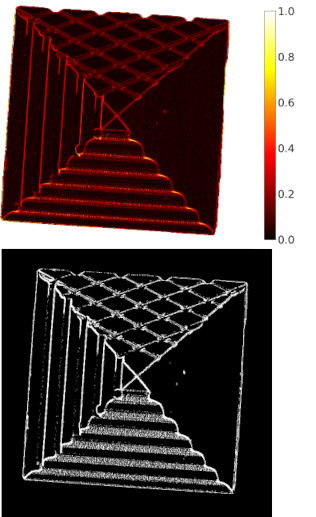}
\caption{}
\label{fig:confidence_and_edges}
\end{subfigure} 
\caption{Our pipeline for performing photometric stereo at the scale of a robot's workspace. \Cref{fig:light_channels} displays the images captured by the camera $C_2$ using each illumination source -- $L_1$ through $L_6$ with the background automatically removed. The top image of \cref{fig:top_channel_and_normal} is the image captured using $L_7$ and the bottom image represents the normals obtained after solving \cref{eq:photometric_loss}. The normal slopes about world's X,Y and Z axes (see \cref{fig:workspace_schematic} for reference) have been mapped to red, green and blue channels respectively. \Cref{fig:pcd_textured_pyramid} presents the heightmaps of the object constructed by integrating the normals in \cref{fig:top_channel_and_normal}, and, \cref{fig:confidence_and_edges} from top to bottom presents our setup's confidence about a pixel being a depth edge and the recovered edge map.}
\label{fig:pipeline_master}
\end{figure*}

\subsection{Modelling the illumination of the workspace}\label{sc:calib_light_table}
We choose to illuminate the robot workspace with low angle of incidence lighting, with approximately parallel light rays, emulating a light source at infinity. This arrangement is also known as grazing illumination in the literature (see e.g. \cite{Johnson2011}). To achieve grazing illumination in practice, we use rectangular shaped light panels larger than our objects and mount the robot so the center of the robot's workspace is approximately equidistant from all our light sources. 
% As the light sources are fixed with respect to the robot's base, we expect to be able to recover effect of the geometric location of the lights on the image captured by the cameras. 
In this section, we describe our approach for mathematically modelling the illumination in the robot's coordinate frame. To recover our illumination model, we capture images from the two cameras ($C_1$ and $C_2$) with only one of each light (($L_{1:6}$ in \cref{fig:workspace_schematic}) on. The calibration object is a matte white hemishperical target with a 4cm radius. \newline
% we capture six images per camera (two cameras) looking at a white hemispherical target of radius 40mm with each of the directional lights ($L_{1:6}$ in \cref{fig:workspace_schematic}) on the table. \newline
\indent Following \cite{Johnson2011}, we choose to model the  illumination seen by a camera using a linear and a quadratic model. The linear illumination model, following \cite{Barsky2003}, is the Lambertian reflectance equation
\begin{align}\label{eq:lambertian_shading}
I_i^k = \frac{\rho}{\pi}\langle\mathbf{n}_k, \mathbf{l}_i \rangle ~ ~ \forall i \in 1...,6
\end{align}
where $I_i^k$ is the intensity of the $k^{\rm{th}}$ pixel, given the $i^{\rm{th}}$ light of $L_{1,...,6}$ is switched on, $\mathbf{n}_k$ is the normal vector at the world point corresponding to the $k^{\rm{th}}$ pixel in the manipulator's coordinate frame (see \cref{fig:workspace_schematic} for reference), $\mathbf{l}_i$ is the direction of the illumination of the $i^{\rm{th}}$ light source and $\rho$ is the albedo of the surface of our target. By design, no three illumination vectors of our approach are co planar -- they are better approximated by vectors on the surface of a frustum, so for each pixel we follow \cite{Barsky2003}, and use the Moore-Penrose operator to invert \cref{eq:lambertian_shading} to get the illumination vectors $\mathbf{l}_i, i \in (1,...,6)$ at the $k^{\rm{th}}$ image pixel. As the albedo is constant on the reflective surface, we include it inside the recovered illumination vectors. 
% \indent We setup the cameras to look approximately vertically down on the hemispherical calibration target and localize the hemisphere in the robot's coordinate frames. This information is then used to generate normal vectors $\mathbf{n}$ in the robot's coordinate frame for each of the pixels imaged by a camera. This observation is independent of the cameras, but we choose to use the images captured with a 16mm lens as it's narrower field of view yields higher resolution image of the workspace. We recall the Lambertian reflectance model
\newline \indent The linear shading model, however, does not capture the effects of ambient illumination (room lights), the effects of lights bouncing off of walls to re-illuminate the scene (e.g. light from illumination channels $L_1,~L_2,~L_3$ bouncing off the wall (see \cref{fig:workspace_schematic}) and re-illuminating the scene), and the errors due to our assumption that we have directional light sources at infinity. To approximately address these artifacts we consider a quadratic shading model derived from a truncated spherical harmonic shading model, which has been shown to be a good approximation of Lambertian reflectance under arbitrary illumination conditions. Researchers have used this model for general purpose rendering (see e.g. \cite{Ramamoorthi2001, Basri2003}), for small scale micro-geometry capture systems (\cite{Johnson2011}) and for recovering shape and illumination from shading of images (e.g. \cite{Barron2014}). We follow the formulation in \cite{Ramamoorthi2001} and can write the quadratic shading model as 
\begin{align}\label{eq:quadratic_shading}
    I_i^k = {\tilde{\mathbf{n}}_k}^T\mathbf{M}_i{\tilde{\mathbf{n}}_k} ~ ~ \forall i \in 1...,6
\end{align}
where, $\tilde{\mathbf{n}}_k = \left[\mathbf{n}_k, 1 \right]^T $ and $\mathbf{M}_i$ is a $4\times4$ symmetric matrix with 9 independent quantities (spherical harmonic lighting coefficients) per illumination source. For each light source, we expand \cref{eq:quadratic_shading} and solve a system of linear equations  to obtain the lighting coefficients. 
\newline \indent Finally, we observe that due to the limited power of the lights and further errors in our assumptions on modelling the lights and the sensitivity of the cameras, the illumination coefficients vary at different parts of the workspace and to take this into account, we take several images of our calibration target (20-25 placements of the target in a workspace of 460mm $\times$ 610mm) and use a bi-quadratic spline interpolation to model the spatial variation of the linear and quadratic illumination coefficients in the robot's workspace. Higher order illumination models were unnecessary given that our objects were Lambertian.

\subsection{Recovering surface normals from images}\label{sc:normal_recovery}
With the linear and quadratic illumination coefficients recovered, and the object reflectances made uniform and known due to the white paint, we expect to be able to obtain the \emph{shape} (world coordinate normals at image pixels in a camera) of an object as seen by the cameras. Although, this should be as simple as evaluating \cref{eq:lambertian_shading} at every pixel to get a initial estimate of the shape and then refining it using \cref{eq:quadratic_shading}, cast shadows and unequal influences of the light sources at every pixel due to the relative location of the object and the lights slightly complicate the shape recovery. In this section we describe our approach for recovering the normals at each of the imaged pixels by reasoning about their illumination or shading. We note that it is known from literature (see e.g. \cite{Tappen2002, Rother2011}) that recovering shadow contours yields better performance in shape from shading problems but we choose to reason locally at every pixel to recover shape. With our choice of large image sizes, narrow field of view lenses, and freedom of locating the camera in the workspace, we can typically get a resolution upwards of 50 pixels/mm$^2$ and can achieve reasonable shape estimates through local reasoning. \newline
% In this section we describe our method to address the aforementioned effects and recover the shape of the object from captured illumination. For this work, we reason about the shape conditioned on the contribution of each of the light sources at every pixel independently and do not reason about the cumulative structures of shadows by explicitly recovering connected-ness information of the shadow pixels. It is known from literature (see e.g. \cite{Tappen2002, Rother2011}) that recovering shadow contours yields better performance in shape from shading problems, but for our case with choice of large image sizes, narrow field of view lenses, and locating the cameras, we could typically get a resolution upwards of 50 pixels/mm$^2$ and could achieve reasonable shape through local reasoning. \newline
\indent To get an initial estimate of whether a pixel is illuminated or shadowed, we compare the intensity of the pixel for the images captured using $L_{1:6}$ with the image captured with $L_7$ (see \cref{fig:light_channels_star}). We generate an initial binary mask for shadowed and illuminated pixels by observing that the pixels illuminated due to a directional light source ($L_{1:6}$) would \emph{almost always} be brighter than the pixels illuminated with an overhead source ($L_7$). With this, we get a binary shadow-illumination vector $\mathbf{w}_k \in \{0,1\}^6$ for all the directional sources ($L_{1:6}$) and augment it to get a binary diagonal shadow-illumination matrix $\mathbf{W}_k = \rm{diag}(\mathbf{w}_k)$. Using $\mathbf{W}$ and following \cite{Barsky2003} we can invert a weighted version of \cref{eq:lambertian_shading} per-pixel to get the initial estimates of \emph{shape} at each pixel as:
\begin{align}\label{eq:shape_linear_estimate}
    \hat{\mathbf{n}}_k = (\mathbf{W}_k\mathbf{L}_{lin}^k)^\dagger\mathbf{W}_k\mathbf{i}_k
\end{align}
where $\mathbf{L}_{lin}^k \in \mathbb{R}^{6\times3}$ is the concatenated illumination matrix at the world location of the k$^{\rm{th}}$ pixel for each of the illumination sources, obtained by concatenating $\mathbf{l}_i \forall i \in 1,...,6$, and $\mathbf{i}_k$ is the vector of all the intensities observed at the k$^{\rm{th}}$ pixel due to $L_{1:6}$. $(\cdot)^\dagger$ is the Moore-Penrose inverse operator. 
\newline \indent However, in practice, the effect of the shadows are not binary and if a pixel is shadowed across more than 3 light channels, which happens often for undercut surface features, \cref{eq:shape_linear_estimate} is not solvable. We frequently encountered this scenario for objects like the textured pyramid, the Buddha, the bunny, and while imaging dough balls (high resolution images on paper website). In these cases, the estimates from \cref{eq:shape_linear_estimate} were incorrect and we assigned shadowed pixels the normals of their nearest valid neighboring pixel. This introduces high local errors in the normal estimates and to address this, along with our motivation for recovering the quadratic shading model leads us to jointly refine the normal estimates $\hat{\mathbf{n}}_k$ and shadow contributions $\mathbf{w}_k$ from \cref{eq:shape_linear_estimate}. To do this, we use the previously estimated $ \hat{\mathbf{n}}_k$ in \cref{eq:quadratic_shading} to estimate the intensities $\hat{I}_i^k, i \in(1,...,6)$ of each pixel conditioned on which light source ($L_{1:6}$) was switched on. We then weigh the intensities with their corresponding shadow weights $\mathbf{w}_k$ and compare the predicted intensities to the observed intensities $I_i^k, ~i\in {1,...,6}, \forall k$ to get a per pixel loss $\ell_k = ||I_i^k - \hat{I}_i||_2$. Finally, we iteratively solve a regularized photometric loss (\cref{eq:photometric_loss}) across all pixels and channels  while updating our hypotheses of per-pixel shadow weights using \cref{eq:coordinate_descent_shadow} at every step. 
\begin{align}\label{eq:photometric_loss}
    \sum_{k,i}(||I_i^k - {w}_k^i\hat{I}_i^k ||_2) + \beta\mathcal{L}_d(\mathbf{n}_k) ~i\in (1,...,6), \forall k 
\end{align}
\begin{align}\label{eq:coordinate_descent_shadow}
    \mathbf{w}_k^{t+1} = \left\{ \begin{tabular}{c}
        $\mathbf{w}_k^{t}$, ~$\rm{if}~ \ell_k^{t+1} \leq \ell_k^{t}$  \\
        $\epsilon \max\{\epsilon, \ell_k^{t+1}\}$ ~ $\rm{if}~ \ell_k^{t+1} > \ell_k^{t}$
    \end{tabular} \right\}
\end{align}
As we treat every pixel individually, we incorporate the intuition that neighboring pixels (or world points) should have similar normals (also known as an integrability constraint in \cite{Horn1970, Johnson2011} or smoothness priors in \cite{Barron2014}) through a Laplacian cost $\beta\mathcal{L}_d(\cdot)$ in \cref{eq:photometric_loss}. The influence of this regularizer can be controlled through the width of the Laplacian filter $d$ and the weight $\beta$. We minimize \cref{eq:photometric_loss} using gradient descent using backtracking line search (\cite{Wright1999}) -- variants of stochastic gradient descent were too unstable for our use case. We also note that $\mathbf{n}_k$ is always calculated in the robot's coordinate frame, as was the case with the illumination model and the calculated normals are independent of the camera's orientation in the manipulator's coordinate frame. Our two step refinement is somewhat  similar to the one described in \cite{Johnson2011}, however, our per pixel inference model along with the differentiability incorporated in the computational structure affords large accelerations with modern tensor libraries and GPGPUs. We discuss further implementation details in  \cref{sc:implementation_details}.

\subsection{Recovering object edges and depth}\label{sc:edge_and_depth_recovery}
\textbf{Recovering edges:} We observe that occlusion of a light source at any point on an object surface depends on the relative location of the light source with respect to the object, so occlusion edges change when the light source location is changed. Consider the image due to light source $L_1$ in \cref{fig:light_channels} -- one can intuitively conclude that the light source is at the bottom left of the object and as we move along the light's path (from bottom left to top right), the sudden change in brightness of the pixels denotes a sharp change of object's visibility along the direction. Indeed, the illumination-shadow ridge is along the image of the ridge where two sloped faces of the tetrahedron meet, and the resulting surface patch falls out of the ``view'' of $L_1$. Similar reasoning applies to all the images captured using the illumination channels $L_{2:6}$ in \cref{fig:light_channels}. This observation was used by \cite{Raskar2004} to perform non-photorealistic stylized rendering of images and by \cite{Liu2012} for detecting edges to localize fabricated parts. In this work, we modified Raskar and colleagues'(\cite{Raskar2004}) pipeline to accommodate illumination sources not co-incident with the camera and use the relative orientation between the camera and illumination sources to calculate edges in an object due to depth changes. \Cref{fig:confidence_and_edges} (top to bottom) shows our confidence (following \cite{Raskar2004}) about whether a pixel is at a depth edge and the edge map obtained by hysteresis thresholding (see e.g. Canny \cite{Canny1986}) the confidence map. We provide more details in \cref{sc:occlusion_edge_details}.
% As shown in \cite{Raskar2004}, conventional edge finding methods (\cite{Canny1986}) would work using the ensemble of images in \cref{fig:light_channels}, our modifications to Raskar et al.\cite{Raskar2004} is better at distinguishing depth edges from edges generated due to surface texture . 
\newline \indent
\textbf{Recovering depth:} With the surface normals from the camera's viewpoint recovered, we can spatially integrate the normals to get a representation of the surface in 3D, as imaged by the camera. This is a classically studied problem in vision known as ``shape from shading'' and there are several solutions to this problem in the literature. We looked at four classical solutions given along the first column of \cref{tab:normal_integration_methods}. We benchmarked them in four aspects: computation speed (speed), robustness to local errors in calculated normals due to shadows, accuracy of surface reconstruction (or absence of strong global smoothing priors) and admissibility of non-axis-aligned arbitrary quadrilaterals image patches. \Cref{tab:normal_integration_methods} summarizes our qualitative findings and for the results in this work, we used the perspective corrected Poisson integration (\cite{Queau2017}). Resulting integrated depth maps can be seen in \cref{fig:pcd_star,fig:pcd_textured_pyramid}, obtained by integrating the normals in \cref{fig:normals_star,fig:top_channel_and_normal} respectively. \newline \indent
However, we should note that, unlike other true depth sensing systems (see \cref{tab:sensor_comparisons}) which directly measure depth  at every pixel of the imaged scene using stereo or time-of-flight sensing, the integrated heightmaps are implicit surfaces that locally have the same normals as the calculated normal map. They will be metrically incorrect unless the whole object is visible from the camera's view port, and the projection scale and the boundary conditions for the integration are not exactly known. We ensured this for the objects presented in \cref{fig:pcd_star,fig:pcd_textured_pyramid}. An object like a cuboid would not work. This limits our approach's applicability as a true depth sensor for manipulation tasks when only one camera view is being used.  Counter-intuitively, this is not a limitation while obtaining depth maps from tactile sensor images (see e.g. Chaudhury et al.\cite{Chaudhury2022} and Johnson et al.\cite{Johnson2011}) where the aim is to only reconstruct the deformed sensor surface - not the object beyond the deformed gel membrane. However, apart from the small sensing footprint, elastomeric tactile sensors fail to capture sharp surface details as they drape over the surface discontinuities. 
\begin{table}[!]    
    \centering
    \begin{footnotesize}
    \begin{tabular}{|c|c|c|c|c|}
    \hline
         Method & Speed & Robust & Accurate & Shape  \\
    \hline
        Variational calc. (\cite{Horn1986,Eberly2008}) & {\color{teal}\bf{\checkmark}} & {\color{red}$\bf{\times}$} & {\color{red}$\bf{\times}$}& {\color{teal}\bf{\checkmark}}\\ 
    % \hline
        Fourier (\cite{Frankot1988})  & {\color{teal}\bf{\checkmark}} & {\color{teal}\bf{\checkmark}}& {\color{red}$\bf{\times}$} &{\color{red}$\bf{\times}$}\\ 
    % \hline
        Least sq. (IRLS) (\cite{Johnson2011, Cao2022bilateral}) & {\color{red}$\bf{\times}$} & {\color{teal}\bf{\checkmark}} & {\color{teal}\bf{\checkmark}} & {\color{teal}\bf{\checkmark}}\\ 
    % \hline
        Poisson Int. (\cite{Queau2017})  & {\color{teal}\bf{\checkmark}}  & {\color{teal}\bf{\checkmark}} & {\color{teal}\bf{\checkmark}} &{\color{teal}\bf{\checkmark}} \\ 
    \hline
    \end{tabular}
\end{footnotesize}      
    \caption{Qualitative comparison of the different normal integration methods evaluated. For the results presented in this work, we used the perspective corrected Poisson integration technique. The techniques which had suitable behavior in our test criteria have been marked with a {\color{teal}\bf{\checkmark}} sign, and unsuitable behavior has been marked with a {\color{red}$\bf{\times}$} sign.}
    \label{tab:normal_integration_methods}
\end{table}

\section{Details of our steps for calculating occlusion edges}\label{sc:occlusion_edge_details}
For detecting occlusion edges we adapt the algorithm described in \cite{Raskar2004} to suite our approach with lights far away from the camera. We first collect the directionally illuminated edges $\mathbf{I}_{1:6}$ corresponding to $L_1$ through $L_6$ and the image $\mathbf{I}_7$ with the overhead light $L_7$ (see \cref{fig:workspace_schematic}). We use the intuition that, given the viewpoint remains fixed, a portion of the scene illuminated with a directional source ($\mathbf{I}_{1:6}$  due to $L_{1:6}$) would be highlighted in contrast to an image due to the overhead light ($\mathbf{I}_7$ due to $L_7$). Following \cite{Raskar2004} we calculate the ratio images in \cref{eq:ratio_images}, making adjustments to avoid numerical errors. 
\begin{equation}\label{eq:ratio_images}
    \mathbf{R}_{1:6} = \dfrac{\mathbf{I}_{1:6}}{\mathbf{I}_7}
\end{equation}
The ratio values in $\mathbf{R}_{1:6}$ will be higher for the areas illuminated with the directional sources ($L_{1:6}$) and lower for areas in shadows, with a distinct transition (from low to high values) along the occlusion edges. To further highlight the individual contributions of the directional illumination, we jointly reason about the transitions in $\mathbf{R}_{1:6}$ conditioned on the direction of the illuminating source $L_{1:6}$ with respect to the camera. For each light source $L_{1:6}$ making an angle $\theta_{1:6}$ with the image axis (camera axes X and Y correspond to the vertical and horizontal image axes respectively), we extract the pixel values in $\mathbf{R}_{1:6}$ with strong transitions along a direction of $\theta_{1:6}$. This gives us a measure of our confidence that a pixel is on an occlusion edge (see top image of \cref{fig:confidence_and_edges}). Finally following \cite{Canny1986}, we apply hysteresis thresholding to the confidence map and obtain a binary map of occlusion edges (bottom image of \cref{fig:confidence_and_edges}). \newline 

\section{Description of our algorithm for vacuum grasping}\label{sc:pickup_details}
Our pipeline for detecting a flat face along an axis can be summarized with the following steps:
\begin{enumerate}
    \item We first obtain the normals and the depth edges of the objects in the robot's workspace. The top insets of \cref{fig:12_mm_grasp_selection_45_deg,fig:12_mm_grasp_selection_flat} denote the normals of the scene and the bottom insets denote the depth edges of the scene. 
    \item For a given picking direction, we randomly generate a set of feasible vacuum gripper orientations centered along the picking direction. For example, if the picking direction is along the $+Z$ axis (see \cref{fig:workspace_schematic}), our samples resemble picking directions that are normally distributed around the vertical axis. For picking along $\pm X$ or $\pm Y$ we adjust the sampling to only admit candidate orientations that avoid collision between the gripper and the table. 
    % However, If the given picking direction is $+Y$, the sampled directions are normally distributed about the vector making $45^\circ$ with both the normal to the table and the $+Y$ axis. As half of the sampled directions about either of $\pm X$ or $\pm Y$ are infeasible, due to the possible interpenetration of the gripper and the table, we use this heuristic to be sample efficient. 
    \item Next, we calculate the probability of each of the pixels in the imaged workspace to be approachable by the set of sampled vacuum gripper configurations. If the normal at a pixel is aligned with the picking direction, it is assigned a high probability of success. This is identical to the histogram back-projection step of the CAMShift algorithm. 
    \item Following this, we identify the largest cluster of feasible pixels using adaptive mean shift clustering. We adapt the scale and the orientation of the kernel as prescribed in \cite{Bradski1998}. This step identifies a candidate zone for our grasp. The top rows of \cref{fig:12_mm_grasp_selection_45_deg,fig:12_mm_grasp_selection_flat}  demonstrate the output of this step, projected onto the normals of the scene calculated using the method described in \cref{sc:normal_recovery}. 
    % We note that the two identified areas are geometrically corresponding due to the calculated normals in both cases being aligned to the robot's frame.
    \item Finally, we score the suitability of executing a vacuum grasp on the selected area by noting that any surface patch with depth edges or surface textures would not be suitable for a vacuum grasp. To do this, we project the selected grasp areas on the scene's edges calculated using the method described in \cref{sc:edge_and_depth_recovery} and generate a score based on the $0^{\rm{th}}$ and $1^{\rm{st}}$ pixel moments that indicate the number of edge pixels and their spread inside the chosen grasp area. Lower scores indicate that the selected patch does not have depth edges. The bottom rows of \cref{fig:12_mm_grasp_selection_45_deg,fig:12_mm_grasp_selection_flat} identify the selected grasp in areas without surface textures projected on the object depth edges calculated by our method in \cref{sc:edge_and_depth_recovery}.
\end{enumerate}
We iteratively apply the above steps along all the directions reachable by our robot, namely $\pm X$, $\pm Y$, and $Z$ for both cameras $C_1$ and $C_2$ (see \cref{fig:workspace_schematic} for reference) and score the detected patches. We identify the highest-scoring patches as flat and ``pickable". \Cref{fig:12_mm_grasp_selection_45_deg,fig:12_mm_grasp_selection_flat} denote the corresponding ``pickable" surfaces oriented along $X$ and $Z$ directions of the workspace, imaged with the robot-mounted camera $C_1$. The grasp areas geometrically correspond across two views imaged by $C_1$ and $C_2$ because the scene normals are calculated with respect to the robot's coordinate frame. 

\section{Description of our algorithm for measuring deformation}\label{sc:card_bending_details}
 We assume that we have a uniformly dense high-quality mesh $\mathcal{M}$ of the object consisting of triangles with aspect ratio close to 1. We also assume that we have knowledge of the pose of $\mathcal{M}$ in the robot's coordinate frame and have access to a differentiable renderer $\mathcal{R}$ (we use PyTorch3D\cite{Ravi2020}) that takes the mesh $\mathcal{M}$, its pose and renders its silhouette and its surface normals in the viewport of two cameras $C_1$ and $C_2$. We clamp the card at one end and push it against a horizontal surface ($X-Y$ plane in \cref{fig:card_bending_schematic}) to induce buckling, and with our vision system, we measure the change in curvature on the object and capture its deformed geometry. From \cref{fig:card_bending_schematic}, we note that the normal $\mathbf{n}_f$ measured by the cameras at the face $f$ of the object mesh $\mathcal{M}$ is dependent on the vertex positions $v_f^i,~i=1,2,3$. Given the initial states of all the vertices in $\mathcal{M}$ we calculate changes in the vertex positions such that the calculated normals at the face $f$ match closely to the imaged normals at the same geometric location. We achieve that through the following steps:
\begin{enumerate}
\item We capture images of the scene using all the lights and calculate the scene normals using the method described in \cref{sc:normal_recovery}. We also capture the silhouettes of the deforming object using our knowledge of the object's pose, geometry and the background. We denote the calculated surface normals and silhouettes of the \emph{scene} as $(\mathbf{N}_S^{C_1}, \mathbf{N}_S^{C_2})$ and $(\mathbf{M}_S^{C_1}, \mathbf{M}_S^{C_2})$ respectively. Equivalent quantities are also \emph{rendered} by the differentiable renderer $\mathcal{R}$ as $(\mathbf{N}_S^{C_1}, \mathbf{N}_R^{C_2})$ and $(\mathbf{M}_R^{C_1}, \mathbf{M}_R^{C_2})$ respectively. 
\item If the states of all the vertices of $\mathcal{M}$ are exactly known, the measured and rendered images should be equivalent. We minimize the  measured difference between the rendered and measured quantities by updating the vertex positions of $\mathcal{M}$.
\item Next, we calculate the difference between the rendered and measured quantities for the data corresponding to $C_1$ as: 

\begin{align}\label{eq:card_bending_loss_c1}
    \ell_{C_1}(\mathcal{M}) = \sum_k\left[ 1 - \langle \mathbf{N}_S^{C_1},\mathbf{N}_R^{C_1}\rangle\right] +
     |\mathbf{M}_S^{C_1} - \mathbf{M}_R^{C_1}|^2  
\end{align} 

for all the $k$ pixels in the camera $C_1$'s viewport. We also obtain a similar difference $\ell_{C_2}$ for $C_2$. We compute a cumulative loss for both the views as 
\begin{align}\label{eq:total_camera_loss}
    \ell(\mathcal{M}) = \alpha \ell_{C_1}(\mathcal{M}) + (1-\alpha) \ell_{C_2}(\mathcal{M})
\end{align}
where $\alpha$ is the fraction of the number of pixels imaging the deforming object in view $C_1$ with the total number of pixels imaging the object across both views $C_1$ and $C_2$.  
\item Minimizing \cref{eq:total_camera_loss} should, in theory, be enough for finding the new locations of the vertices of $\mathcal{M}$, but in practice, local measurements and the nature of gradient descent do not generate a smooth and physically plausible
 mesh through the gradient updates. To address this, we follow \cite{Luan2021} and add two regularizers based on physics -- $L_{Lap.}$: the mesh Laplacian regularizer which encourages geometrically smooth updates to the mesh vertices, $L_{edge}$: the mesh edge length regularizer that promotes mesh vertex updates that keep the aspect ratios of the mesh elements close to 1, to augment our loss in \cref{eq:total_camera_loss}. 
\item Finally, we formulate our objective function as:
\begin{align}\label{eq:bending_obj_fn}
\min_{v_i} \ell(\mathcal{M}) + L_{Lap.}(\mathcal{M}) + L_{edge}(\mathcal{M}) ~ \forall v_i \in \mathcal{M}
\end{align}
We use gradient descent with backtracking line search to minimize \cref{eq:bending_obj_fn} above. 
\end{enumerate}

\section{Description of our pose estimation pipelines}\label{sc:pose_estimation_details}
\textbf{If a 3D model of an object is available}, we define the pose estimation problem as finding the rigid transform $\mathbf{T}_{base}^{obj}$ which aligns the captured image of the object to a rendered equivalent given the camera parameters $\mathbf{K}_i$ and camera poses $\mathbf{T}_{base}^{C_i}$ for cameras $C_1$ \textit{or} $C_2$. In addition to the 3D model $\mathcal{M}$ of the object being available to us, we also assume that we have access to a differentiable renderer $\mathcal{R}$ that can render the 3D model $\mathcal{M}$ given $\mathbf{K}_i$,  $\mathbf{T}_{base}^{C_i}$ and  $\mathbf{T}_{base}^{C_i}$. We use PyTorch3D\cite{Ravi2020} for our work. Using the method described in \cref{sc:normal_recovery,sc:edge_and_depth_recovery} we process our images captured by a camera -- say $C_1$ of a view-port of $w\times h$ pixels, to obtain the \emph{scene} normal map $\mathbf{N}_S\in \mathbb{R}^{w\times h\times3}$, the \emph{scene} depth edges $\mathbf{E}_S \in \mathbb{R}^{w\times h\times1}$ and the object silhouette from the \emph{scene} $\mathbf{M}_S \in \mathbb{R}^{w\times h\times1}$.  \newline
\indent With $\mathcal{R}(\mathcal{M}, \mathbf{T}_{base}^{obj}, \mathbf{T}_{base}^{C_1}, \mathbf{K}_1)$ we also \emph{render} equivalent normal, depth edge and silhouette images $\mathbf{N}_R, ~\mathbf{E}_R$ and $\mathbf{M}_R$. Given that the object pose $\mathbf{T}_{base}^{obj}$ has been correctly estimated, the rendered and the scene images for normals, depth edges and object silhouettes should exactly match. In the following steps, we describe our steps for aligning the scene and rendered data. For our case, $\mathbf{T}_{base}^{obj}$ is parameterized by the position of the object along the X and Y axes and a rotation $\theta$ about Z axis.
\begin{enumerate}
    \item To generate initial estimates of poses, we find correspondences between the measured surface normals $\mathbf{N}_S$ and rendered surface normals $\mathbf{N}_R$ for a given set of $\theta \in [0^\circ, 180^\circ] $. To do this, we identify object depth corners in $\mathbf{E}_S$ and look for patches in $\mathbf{N}_R$. The first inset of \cref{fig:model_based_pose_est} shows some corresponding depth edge corners overlaid on $\mathbf{N}_S$ and $\mathbf{N}_R$ for a particular $\theta$. We find the matches by looking at $80\times80$ windows around the depth corners in $\mathbf{N}_S$ and match them to $\mathbf{N}_R$ for a particular $\theta$ using 3D cross-correlation. These matches, along with $\mathbf{K}_1$ can be used to compute the essential matrix between $\mathbf{N}_S$ and $\mathbf{N}_R$. We then decompose the essential matrices for all $\theta$ and identify the decomposition which produces a rigid transform closest to identity rotation. The corresponding $\theta_i$ is our initial guess for the object's orientation. 
    \item Next, we align the silhouettes $\mathbf{M}_S$ and $\mathbf{M}_R|(x,y,\theta_i)$ using a pixel-wise squared $L_2$ cost summed over all the $k$ pixels in the $w\times h$ simulated and real camera view ports
    \begin{align}\label{eq:silhouette_align}
        \min_{x,y,\theta} \sum_k \left| \mathbf{M}_S^k - \mathbf{M}_R^k|_{(x,y,\theta)} \right|_2^2
    \end{align}
    \item Following which, we align the rendered and measured surface normals $\mathbf{N}_R$ and $\mathbf{N}_S$ respectively about 4 pyramid levels. To do this we generate the normal images  $^i\mathbf{N}_S, ^i\mathbf{N}_R,~i\in(1,2,3,4)$ across different scales and calculate the cosine similarity between the $k$ corresponding pixels inside the corresponding scaled silhouettes $^i\mathbf{M}_S$ :
    \begin{align}\label{eq:normal_align}
        \min_{x,y,\theta}\sum_{k \in ^i\mathbf{M}_S}\left[1 - \langle ^i\mathbf{N}_S^k, ^i\mathbf{N}_R^k|_{(x,y,\theta)} \rangle \right] \forall i
    \end{align}
    where $i=1$ is the lowest pyramid level.
    \item Finally, following Chaudhury et al.\cite{Chaudhury2022} we align the rendered and measured depth edge images $\mathbf{E}_S$ and $\mathbf{E}_R$ by minimizing the mean squared error between the Euclidean distance transforms (EDT) of images $\mathbf{E}_S$ and $\mathbf{E}_R$. Using the definition of Euclidean distance transform from \cite{Felzenszwalb2012}, we formulate our dense edge alignment cost as
    \begin{align}\label{eq:edge_align}
    \min_{x,y,\theta} \sum_k \left| \rm{EDT}(\mathbf{E}_S) - \rm{EDT}(\mathbf{E}_R|_{(x,y,\theta)}) \right|_2^2
    \end{align}
    for all the $k$ pixels in the $w\times h$ simulated and real camera view ports.
\end{enumerate}
\indent If a \textbf{3D model of the object is unavailable}, we define the pose estimation problem as finding the rigid transform $\mathbf{T}$ that aligns the two 3D representations captured by our system as the object moves.   
\begin{enumerate}
\item We first image the object with our camera -- the leftmost inset of \cref{fig:model_free_pose_est} shows the image of a folding knife captured by $C_1$ with $L_7$ illuminating the scene.
\item Next, we calculate the normals of the scene and identify the object depth edges -- these are shown as the second and third insets of \cref{fig:model_free_pose_est}. These normals are only used to generate a 3D representation of the object.
\item Following that we calculate the point cloud of the surface of the object and also calculate per point features -- we use Fast Point Feature Histograms (FPFH) \cite{Rusu2009}. The fourth inset of \cref{fig:model_free_pose_est} shows the point cloud of the object with the salient points colored. We note that these points roughly indicate the depth discontinuities by referring between the third and fourth insets of \cref{fig:model_free_pose_est}.
\item We repeat the steps above for the data obtained at the new pose of the object. 
\item Finally, we use the FPFH in a robust point feature based pose estimation pipeline (we use FGR \cite{Zhou2016}) to generate initial pose estimates and then use point-to-plane ICP \cite{Rusinkiewicz2001} to solve for the change in pose of the object between the two measurements. The last two insets of \cref{fig:model_free_pose_est} shows the two initial measurements of the object on top and the latter measurement registered to the former measurement on the bottom. Point cloud normals need to be re-calculated in the object's frame for the ICP based refinement step.
\end{enumerate}

\section{Implementation details and hyperparameters}\label{sc:implementation_details}

In this work, we attempt to capture and process all the captured data online for the demonstrated perception tasks. To do this, we implemented almost all of the pipelines with strong GPU support. The algorithms for solving the optimization problems associated with  normal estimation (\cref{eq:photometric_loss}), bending estimation (\cref{eq:bending_obj_fn}) and all the alignment costs (\cref{eq:silhouette_align,eq:normal_align,eq:edge_align}) have been implemented with custom GPU back-ends, which led to massive speedup even with a resolution of 50px/mm$^2$. This lets us perform all the tasks in less than 30 seconds time per task per object at full resolution. 
A breakdown of the time take by each step is shown in \cref{tab:time_breakdown}. 
\begin{table}
 \begin{footnotesize}
    \centering
    \begin{tabular}{|c|c|c|}
    \hline
         Step &  Processing + [overhead]  \\
         \hline          
          Capture (\cref{fig:light_channels_star}) &  0.20 (per cam.) + [5] \\ 
         % \hline
       Normal (eq. \ref{eq:lambertian_shading},\ref{eq:quadratic_shading}) & 3.5 ($1000^{2}$ px) \\ 
         % \hline
       Depth (\cref{sc:edge_and_depth_recovery}) & 0.15 ($1000^{2}$ px) \\ 
         % \hline
    Edges (\cref{sc:edge_and_depth_recovery}) & 0.05 ($1000^{2}$ px) \\ 
         % \hline
    Pickup (\cref{fig:12_mm_grasp_selection_45_deg}) & 0.5 ($1000^{2}$ px) \\ 
         % \hline
    Pickup (robot) & 20 (pick and drop) \\ 
         % \hline
    Bending (\cref{eq:bending_obj_fn}) & 15 (1500 vert, 300 iter)) + [5] \\ 
         % \hline
    pose est. (eq. \ref{eq:silhouette_align}, \ref{eq:normal_align}, \ref{eq:edge_align}) & 12.5 - 15.5 (150 iter/eq.) + [5] \\ 
         \hline
    \end{tabular}
    \caption{Approximate time breakdown of various steps in seconds for processing a $1000^{2}$ px image}
    \label{tab:time_breakdown}
     \end{footnotesize}
\end{table}

We also discuss our major design decisions and the observed effects of important hyperparameters of the system we discovered during implementing the system below.

\subsection{Practical considerations in design}\label{sc:practical_considerations}

We implemented the system demonstrated in the work on a robot table of $1.5\rm{m} \times 1.5\rm{m}$, The black mat of size ($460\rm{mm} \times 610\rm{mm}$) in \cref{fig:workspace_schematic} denotes the dexterous workspace of the robot. These dimensions governed the placement of the light sources, their power, and some of the camera hyper-parameters (exposure, f-stops) for our experiments. To achieve good illumination of the workspace (see \cref{sc:hyper_params}, item 1), we observed that a 45W white light source was sufficient when placed 500mm away from the objects. So we placed six light sources in an approximately hexagonal pattern around the center of the workspace with a radius of approximately 500mm. Theoretically, as long as there is a measurable effect of a directional light source on the image of the object, our approach would work -- however the more pronounced the effect, the less is the noise in measuring it with a camera. With the constraints above, and our choice of hyperparameters discussed in \cref{sc:hyper_params}, we could obtain a reasonable performance with our system.
\newline
As the lights $L_1$ through $L_6$ (\cref{fig:workspace_schematic,fig:pipeline_master}) were used as grazing directional light sources they were oriented towards the center of the workspace. The exact orientation of the light panels were not measured as those were implicitly recovered by our illumination models discussed in \cref{sc:calib_light_table}. However, through our fixtures, we ensured that the light sources stationary during the experiment. Light $L_7$ was solely used to estimate the shadows and highlights when the scene was illuminated by $L_1$ through $L_6$, and an illumination model was not recovered from it. We experimented with using $L_7$ as another source (by using seven light sources instead of six in \cref{eq:lambertian_shading,eq:quadratic_shading}) and the quality of the normal measurements degraded significantly because of the poorer initial estimates of the shadow-illumination matrix $\mathbf{W}_k$ in \cref{eq:shape_linear_estimate} and a resulting poorer performance of \cref{eq:photometric_loss,eq:coordinate_descent_shadow}. Our two step illumination model requires at least three light sources to illuminate every portion of the scene for \cref{eq:shape_linear_estimate} to be valid. Except simpler shapes (like a hemisphere or a pyramid) a minimum of three light sources are not guaranteed to illuminate 
all the portions on the surface, unless they are at a higher angle of incidence to the surface. However, a higher angle of incidence does not accentuate the depth discontinuities as much as low angle of incidence illumination, which was a requirement for our case. These constraints led us to choose six illumination sources on the table. We observed from initial experiments that fewer sources performed poorer -- especially for more complex shapes and, inspired by small-scale photometric stereo sensors (\cite{GelSight,GelSightMini,Yuan2017gelsight,Chaudhury2022}) we decided to have six illumination sources. 
\subsection{Hyperparameters and their effects}\label{sc:hyper_params}
The performance of our approach is dependent on some crucial hyper-parameters. We note the important hyper-parameters below: 
\begin{itemize}
    \item {Gains, exposure times and lens f-stop numbers are critical hyperparameters for capturing the effects of directional illumination on the scene consistently. For all our experiments, in addition to setting the sensor gamma to 1, we set the sensor gains and black levels to zero and turn off automatic exposure and gain settings. We adjust the lens f-stops and exposure times so that the brightest spot in the image under all illumination conditions is between 75\% to 85\% of the maximum brightness (\texttt{uint16\_max} 65535). For our setup, an f-stop of 1.6 on the 12mm focal length lens and 1.2 on the 16mm focal length lens and an exposure time of 2.5 milliseconds worked well. These parameters are also dependent on the color and power of the ambient illumination of the room, distance between the light source and the scene, reflectivity of the ceiling and walls, and the color of the wall. }

    \item We used backtracking line search for all the gradient descent steps (\cref{eq:quadratic_shading,eq:bending_obj_fn,eq:silhouette_align,eq:normal_align,eq:edge_align}) in this work. The initial learning rates were 0.1, with a c value of $10^{-5}$ and an annealing factor of $0.9$. We terminated the gradient descent when the step size was lower than $10^{-7}$ or 150 gradient descent steps have been completed. 

    \item For the pickup tasks, we selected the edge pixel score threshold by setting it to a low value  as we only grasped geometrically flat patches. {The hyperparameter value selected patches with less than 3--5\% of the pixels labeled as edges.}

    \item For estimating bending deformation, we found that locating the cameras ($C_1$ and $C_2$ in \cref{fig:card_bending_schematic}) such that both yield similar sized images gave us the best results while estimating the deformation of the card. For all our experiments in \cref{fig:card_bending_plots}, we made sure that the value of $\alpha$ in \cref{eq:total_camera_loss} was in between 0.45 to 0.55.

    \item Finally, for the global registration step in \cref{fig:model_free_pose_est}, we used a FPFH size of 35, and a voxel downsampling factor of 1.5 in the implementation available on Open3D (\cite{Zhou2018open3d}).
\end{itemize}

%%%%%%%%% REFERENCES
{\small
\bibliographystyle{ieee_fullname}
\bibliography{paper_bib}
}

\end{document}